\documentclass[opre]{informs3-arxiv}

\twocolumn

\SingleSpacedXI

\usepackage{endnotes}
\let\footnote=\endnote

\usepackage{natbib}
 \bibpunct[, ]{(}{)}{,}{a}{}{,}%
 \def\bibfont{\small}%
\TheoremsNumberedThrough     
\ECRepeatTheorems

\EquationsNumberedThrough    

\newcommand{\mv}[1]{\textcolor{black}{#1}}

\usepackage{comment}
\usepackage{graphicx}
\usepackage{algorithm,algpseudocode}
\numberwithin{equation}{section}
\theoremstyle{plain}
\newtheorem{thm}{Theorem}[section]
\newtheorem{lm}{Lemma}[section]
\newtheorem{cor}{Corollary}[section]
\newcommand{\mat}[1]{\mathbf #1}
\newcommand{\reals}{\mathrm{I\! R}}
\newcommand{\ind}{\mathrm{1\!\! I}}
\newcommand{\pa}{\vec{w}}
\newcommand{\pae}{w}
\newcommand{\bo}{\omega}
\newcommand{\dM}{d(\vec{M})}
\newcommand{\aM}{a(\vec{M})}
\newcommand{\win}{d}
\usepackage{bm}
\usepackage{amsmath,amssymb,amsfonts}
\usepackage{multirow}
\renewcommand{\vec}[1]{{\mathbf #1}}
\usepackage{mathpazo}

\begin{document}


\RUNAUTHOR{Vojnovic, Yun and Zhou}

\RUNTITLE{Accelerated MM Algorithms for Ranking Scores Inference from Comparison Data}

\TITLE{Accelerated MM Algorithms for Ranking Scores Inference from Comparison Data}

\ARTICLEAUTHORS{%
\AUTHOR{Milan Vojnovic}
\AFF{Department of Statistics, London School of Economics, London, United Kingdom, WC2A 2AE, \EMAIL{m.vojnovic@lse.ac.uk}} 
\AUTHOR{Se-Young Yun}
\AFF{Graduate School of AI, KAIST, Daejeon, South Korea, 34141, \EMAIL{yunseyoung@kaist.ac.kr}}
\AUTHOR{Kaifang Zhou}
\AFF{Department of Statistics, London School of Economics, London, United Kingdom, WC2A 2AE, \EMAIL{k.zhou3@lse.ac.uk}}
} 

\ABSTRACT{%
The problem of assigning ranking scores to items based on observed comparison data, e.g., paired comparisons, choice, and full ranking outcomes, has been of continued interest in a wide range of applications, including information search, aggregation of social opinions, electronic commerce, online gaming platforms, and more recently, evaluation of machine learning algorithms. The key problem is to compute ranking scores, which are of interest for quantifying the strength of skills, relevancies or preferences, and prediction of ranking outcomes when ranking scores are estimates of parameters of a statistical model of ranking outcomes. One of the most popular statistical models of ranking outcomes is the Bradley-Terry model for paired comparisons (equivalent to multinomial logit model), and its extensions to choice and full ranking outcomes. The problem of computing ranking scores under the Bradley-Terry models amounts to estimation of model parameters. 

In this paper, we study a popular method for inference of the Bradley-Terry model parameters, namely the MM algorithm, for maximum likelihood estimation and maximum a posteriori probability estimation. This class of models includes the Bradley-Terry model of paired comparisons, the Rao-Kupper model of paired comparisons allowing for tie outcomes, the Luce choice model, and the Plackett-Luce ranking model. We establish tight characterizations of the convergence rate for the MM algorithm, and show that it is essentially equivalent to that of a gradient descent algorithm. For the maximum likelihood estimation, the convergence is shown to be linear with the rate crucially determined by the algebraic connectivity of the matrix of item pair co-occurrences in observed comparison data. For the Bayesian inference, the convergence rate is also shown to be linear, with the rate determined by a parameter of the prior distribution in a way that can make the convergence arbitrarily slow for small values of this parameter. We propose a simple modification of the classical MM algorithm that avoids the observed slow convergence issue and accelerates the convergence. The key component of the accelerated MM algorithm is a parameter rescaling performed at each iteration step that is carefully chosen based on theoretical analysis and characterisation of the convergence rate. 

Our experimental results, performed on both synthetic and real-world data, demonstrate the identified slow convergence issue of the classic MM algorithm, and show that significant efficiency gains can be obtained by our new proposed method. 
}%



\maketitle

%


\section{Introduction}

Rank aggregation is an important task that arises in a wide-range of applications, including recommender systems, information retrieval, online gaming, sport competitions, and evaluation of machine learning algorithms. Given a set of items, the goal is to infer ranking scores of items or an ordering of items based on observed data that contains partial orderings of items. A common scenario is that of paired comparisons, e.g., player A defeats player B in a game, product A is preferred over product B by a user, and machine learning algorithm A outperforms machine learning algorithm B in an evaluation. In such scenarios, a common goal is not only to compute an aggregate ranking of items, but also to compute ranking scores, which represent strengths of individual items. Such ranking scores are used for predicting outcomes of future ranking outcomes, such as predicting outcomes of matches in online games and sport contests, and predicting preference of users in product shopping or movie watching scenarios, among others. Note that, importantly, observations are not restricted to paired comparisons, but may also include other types of comparison data, such as choice (e.g., product A chosen from a set of two or more products) or full ranking (e.g., a ranking list of players or teams participating in a competition).

In this paper, our goals are (a) to shed light on the efficiency of one of the most popular iterative optimization methods for inference of ranking scores, namely the MM algorithm, where ranking scores correspond to parameter estimates of popular Bradley-Terry family of models, and (b) to propose an accelerated MM algorithm that resolves a slow convergence issue found to hold for a classic MM algorithm.  

\paragraph{\bf Related work} Statistical models of ranking data play an important role in a wide range of applications, including learning to rank in information retrieval (\cite{BRL07,Li11}), skill rating in sport games (\cite{E78}), online gaming platforms (\cite{TrueSkill07}), and evaluation of machine learning algorithms by comparing them with each other (\cite{Eval18}). 

A common class of statistical models of ranking data are \emph{generalized Bradley-Terry models}, which accommodate paired comparisons with win-lose outcomes (\cite{Z29, BT52,B54}), paired comparisons with win-lose-draw outcomes (\cite{RK67}), choices from comparison sets of two or more items, e.g., Luce choice model (\cite{L59}), full ranking outcomes for comparison sets of two or more items, e.g., Plackett-Luce ranking model (\cite{P75}), as well as group comparisons (\cite{Huang06,HLW08}). These models can be derived from suitably defined latent variable models, where items are associated with independent latent performance random variables, which is in the spirit of the well-known Thurstone model of comparative judgment (\cite{Thur27}).


An iterative optimization algorithm for the maximum likelihood (ML) parameter estimation (MLE) of the Bradley-Terry model has been known since the original work of \cite{Z29}. \cite{LHY00} showed that this algorithm belongs to the class of MM optimization algorithms. Here MM refers to either minorize-maximization or majorize-minimization, depending on whether the optimization problem is maximization or minimization of an objective function. \cite{L16} provided a book on MM algorithms and \cite{Hunter04tut} provided a tutorial. \cite{M15} established some convergence results for incremental MM algorithms. 

In a seminal paper, \cite{hunter04} derived MM algorithms for generalized Bradley-Terry models as well as sufficient conditions for their convergence to ML estimators using the framework of MM optimization algorithms. For the Bradley-Terry model of paired comparisons, a necessary and sufficient condition for the existence of a ML estimator is that the directed graph whose vertices correspond to items and edges represent outcomes of paired comparisons is connected. In other words, the set of items cannot be partitioned in two sets such that none of the items in one partition won against an item in other partition. 

A Bayesian inference method for generalized Bradley-Terry models was proposed by \cite{Caron12}, showing that classical MM algorithms can be reinterpreted as special instances of Expectation-Maximization (EM) algorithms associated with suitably defined latent variables and proposed some original extensions. This amounts to MM algorithms for maximum a posteriori probability (MAP) parameter estimation, for a specific family of prior distributions. This prior distribution is a product-form distribution with Gamma$(\alpha,\beta)$ marginal distributions, where $\alpha \geq 1$ is the \emph{shape} parameter and $\beta > 0$ is the \emph{rate} parameter. Importantly, unlike to the ML estimation, the MAP estimate is always guaranteed to exist, for any given observation data.  

\mv{Algorithms for fitting Bradley-Terry model parameters are implemented in open source software packages, including BradleyTerry2 (\cite{TF12}), BradleyTerryScalable (\cite{BTScalable}), and Choix (\cite{LM18}). The first package uses a Fisher scoring algorithm (a second-order optimization method), while the latter two use MM algorithms (a first-order optimization method). First-order methods are generally preferred over second-order methods for fitting high-dimensional models using large training datasets.}

While the conditions for convergence of MM algorithms for generalized Bradley-Terry models are well understood, to the best of our knowledge, not much is known about their \emph{convergence rates} for either ML or MAP estimation. In this paper, we close this gap by providing tight characterizations of convergence rates. Our results identify key properties of input data that determine the convergence rate, and in the case of MAP estimation, how the convergence rate depends on prior distribution parameters. Our results show that MM algorithms, commonly used for MAP estimation for generalized Bradley-Terry models, can have a slow convergence for some prior distributions. 

Recent research on statistical models of paired comparisons focused on characterization of the accuracy of parameter estimators and development of new, scalable parameter estimation methods, e.g., \cite{GS09,WJJ13,hajek14,RA14,CS15,SBBPRW16,Vojnovic16b,KO16,Negahban17,borkar2016randomized,CFMW19}. \mv{Note that the question about statistical estimation accuracy and computation complexity tradeoff is out of the scope of our paper, and this was studied in the above cited papers. The focus of our work is on \emph{convergence properties} of \emph{first-order} iterative optimization methods for parameter estimation of Bradley-Terry models. Here "first-order" refers to optimization methods that are restricted to value oracle access to gradients of the optimization objective function, thus not allowing access to second-order properties such as values of the Hessian matrix. Specifically, we are interested in convergence properties of first-order methods for ML and MAP estimation objectives.} It is noteworthy that some recently proposed algorithms show empirically faster convergence rate than MM, e.g., \cite{Negahban17,maystre2015fast,agarwal2018accelerated}, but it is hard to apply them for the MAP estimation objective. We thus restrict our attention to MM and gradient descent algorithms which are able to solve both MLE and MAP optimization problems. 

A preliminary version of our paper was published in \cite{VYZ2020}, which contains results on convergence rates of gradient descent and MM algorithms (Section~\ref{sec:conv-bt}). In the present paper, we extend our prior work by proposing new accelerated algorithms and establishing their theoretical guarantees (Section~\ref{sec:acc}) as well as demonstrating their efficiency through numerical evaluations (Section~\ref{sec:num}).

\paragraph{\bf Summary of our contributions} 
We present tight characterizations of the rate of convergence of gradient descent and MM algorithms for ML and MAP estimation for generalized Bradley-Terry models.  Our results show that both gradient descent and MM algorithms have linear convergence with convergence rates differing only in constant factors. \mv{An iterative optimization algorithm that has linear convergence is generally considered to be fast \emph{in the space of first-order optimization algorithms}, and many first-order algorithms cannot guarantee a linear convergence. For example, standard stochastic gradient descent algorithm is known to have sub-linear convergence, see, e.g., \cite{Bubeck15}.} We provide explicit bounds on convergence rates that provide insights into which properties of observed comparison data play a key role for the rate of convergence.  

Specifically, we show that the rate of convergence critically depends on certain properties of the matrix of item pair co-occurrences, $\vec{M}$, of input comparison data. We found that two key properties are: (a) maximum number of paired comparisons per item (denoted as $\dM$) and (b) the algebraic connectivity of matrix $\vec{M}$ (denoted as $\aM$). Intuitively, $\aM$ quantifies how well is the graph of paired comparisons connected. Here $\aM$ is the Fiedler value (eigenvalue), cf. \cite{F73}, defined as the second smallest eigenvalue of the Laplacian matrix $\mat{L}_{\mat{M}} = \vec{D}_{\vec{M}} - \vec{M}$, where $\vec{D}_{\vec{M}}$ is the diagonal matrix with diagonal elements equal to the row sums of $\vec{M}$. 
The Fiedler value of a matrix of paired comparison counts is known to play a key role in determining the MLE accuracy, e.g.,  \cite{hajek14,SBBPRW16,khetan2016data,Vojnovic16b,Negahban17}. These works characterized the number of samples needed to estimate the true parameter vector within a statistical estimation error tolerance. This is different from the problem of characterizing the number of iterations needed for an iterative optimization algorithm to compute a ML or a MAP parameter estimate satisfying an error tolerance condition, which is studied in this paper. 

Our results reveal the following facts about convergence time, defined as the number of iterations that an iterative optimization algorithm takes to reach the value of the underlying objective function within a given error tolerance parameter $\epsilon > 0$ of the optimum value.  

For the ML objective, we show that the convergence time satisfies
\begin{equation*}
T^{\mathrm{ML}} = O\left(\frac{\dM}{\aM}\log\left(\frac{1}{\epsilon}\right)\right)
\label{equ:ctmle}
\end{equation*}
which reveals that the rate of convergence critically depends on the connectivity of the graph of paired comparisons in observed data.

For the MAP estimation, we show that the convergence time satisfies
\begin{equation*}
T^{\mathrm{MAP}} = O\left(\left(\frac{\dM}{\beta}+1\right)\log\left(\frac{1}{\epsilon}\right)\right)
\label{equ:ctmap}
\end{equation*}
where, recall, $\beta > 0$ is the rate parameter of the Gamma prior distribution. This bound is shown to be tight for some input data instances. We observe that the convergence time for the MAP estimation problem can be arbitrarily large for small enough parameter $\beta$, where small values of parameter $\beta$ correspond to less informative prior distributions. 

Our results identify a slow rate of convergence issue for gradient descent and MM algorithms for the MAP estimation problem. While the MAP estimation alleviates the issue of the non-existence of a ML estimator when the graph of paired comparisons is disconnected, it can have a much slower convergence than ML when the graph of paired comparisons is connected. Perhaps surprisingly, the rate of convergence has a discontinuity at $\beta = 0$, in the sense that for $\alpha = 1$ and $\beta = 0$, the MM algorithm for the MAP estimation corresponds to the classic MM algorithm for ML estimation, and in this case, the convergence bound (\ref{equ:ctmle}) holds, while for the MAP estimation, the convergence time grows arbitrarily large as $\beta$ approaches $0$ from above.

We propose an acceleration method for the MAP estimation objective, with convergence time bounded as follows
\begin{equation*}
T^{\mathrm{MAP}}_{\mathrm{Acc}} = O\left(\min\left\{\frac{\dM}{\aM}, \frac{\dM}{\beta}\right\}\log\left(\frac{1}{\epsilon}\right)\right).
\label{equ:ctmapacc}
\end{equation*}
This acceleration method resolves the slow convergence issue of classic MM algorithm for the MAP estimation for generalized Bradley-Terry models. This accelerated method does not have a discontinuity at $\beta = 0$ with respect to the rate of convergence: as $\beta$ approaches $0$ from above, the convergence time bound corresponds to that of the MM algorithm for ML estimation. The acceleration method \mv{normalizes the parameter vector estimate in each iteration of the gradient descent or MM algorithm using a particular normalization that ensures (a) that the value of the objective function is non decreasing along the sequence of estimated parameter vectors and (b) that the objective function satisfies certain smoothness and strong convexity properties that ensure high convergence rate. This amounts to a slight modification of the classical MM algorithm that resolves the identified convergence issue. This acceleration method is derived by using a theoretical framework that may be of general interest. This framework can be applied to different statistical models of ranking data and prior distributions for Bayesian inference of parameters of these models.} 

We present numerical evaluation of the convergence time of different iterative optimization algorithms using input data comparisons from a collection of real-world datasets. These results demonstrate the extent of the slow convergence issue of the existing MM algorithm for MAP estimation and show significant speed ups achieved by our accelerated MM algorithm.  

Our theoretical results are established by using the framework of convex optimization analysis and spectral theory of matrices. In particular, the convergence rate bounds are obtained by using concepts of smooth and strongly convex functions. We derive accelerated iterative optimization algorithms based on a general approach that may be of independent interest. This approach transforms the parameter estimator in each iteration so that certain conditions are preserved for the gradient vector and the Hessian matrix of the objective function. 
For generalized Bradley-Terry models, this transformation turns out to be simple, yielding a practical algorithm.

\paragraph{\bf Organization of the paper} In Section~\ref{sec:pre}, we present problem formulation and some background material. Section~\ref{sec:conv-bt} contains our main results on characterization of convergence rates of gradient descent and MM algorithms; for simplicity of exposition, we focus only on the Bradley-Terry model of paired comparisons. Section~\ref{sec:acc} presents our accelerated algorithms for MAP estimation. Section~\ref{sec:num} contains our numerical results. We conclude in Section~\ref{sec:disc}. Section~\ref{sec:pro} contains all our proofs, additional discussions, and extensions to generalized Bradley-Terry models.

\section{Problem formulation} 
\label{sec:pre} 

According to the \emph{Bradley-Terry model of paired comparisons} with win-lose outcomes, each comparison of items $i$ and $j$ has an independent outcome: either $i$ wins against $j$ ($i\succ j$) or $j$ wins against $i$ ($j\succ i$). The distribution of outcomes is given by  
\begin{equation}
\Pr[i\succ j] = \frac{\theta_i}{\theta_i + \theta_j}
\label{equ:pij}
\end{equation}
where $\theta = (\theta_1, \theta_2,\ldots,\theta_n)^\top \in \reals_+^n$ is the parameter vector. 
The Bradley-Terry model of paired comparisons was studied by many, e.g., \cite{Ford57,D56,D60,SY99} and is covered by classic books on categorical data analysis, e.g., \cite{agresti02}. 

We will sometimes use the parametrization $\theta_i = e^{\pae_i}$ when it is simpler to express an equation or when we want to make a connection with the literature using this parametrization. Using parameterization $\pa = (w_1, w_2, \ldots, w_n)^\top \in \reals^n$, we have
$$
\Pr[i\succ j] = \frac{e^{w_i}}{e^{w_i} + e^{w_j}}.
$$

All our convergence results are for the model with parameter $\pa$. 
The Bradley-Terry type models for paired comparisons with ties, choice, and full ranking outcomes, we refer to as \emph{generalized Bradley-Terry models}, are defined in Section~\ref{sec:genBT-app}. Our results apply to all these different models. In the main body of the paper, we focus only on the Bradley-Terry model for paired comparisons in order to keep the presentation simple.   

\paragraph{\bf Maximum likelihood estimation} The maximum likelihood parameter estimation problem corresponds to finding $\pa^\star$ that solves the following optimisation problem:
\begin{equation}
\max_{\pa \in \reals^n}\ell(\pa)
\label{equ:optmle}
\end{equation}
where $\ell(\pa)$ is the log-likelihood function,
\begin{equation}
\ell(\pa) = \sum_{i=1}^n \sum_{j\neq i} \win_{i,j} \left(\pae_i - \log\left(e^{\pae_i} + e^{\pae_j}\right)\right)
\label{equ:btloglik}
\end{equation}
with $\win_{i,j}$ denoting the number of observed paired comparisons such that $i\succ j$.

The maximum likelihood optimisation problem (\ref{equ:optmle}) is a convex optimization problem. Note, however, that the objective function is not a strictly concave function as adding a common constant to each element of the parameter vector keeps the value of the objective function unchanged. 

\paragraph{\bf MAP estimation problem} An alternative objective is obtained by using a \emph{Bayesian inference framework}, which amounts to finding a maximum a posteriori estimate of the parameter vector under a given prior distribution. We consider the Bayesian method introduced by \cite{Caron12}, which assumes the prior distribution to be of product-form with marginal distributions such that $\theta_i (= e^{\pae_i})$ has a \mv{Gamma($\alpha, \beta$) distribution where $\alpha \geq 1$ is the \emph{shape} parameter and $\beta > 0$ is the \emph{rate} parameter. Note that $\alpha$ and $\beta$ affect the \emph{scale} of the parameter vector as with respect to the Gamma($\alpha, \beta$) prior distribution, $\theta_i$ has the expected value and the mode equal to $\alpha/\beta$ and $(\alpha-1)/\beta$, respectively. For any fixed $\alpha \geq 1$, the density of Gamma($\alpha, \beta$) distribution becomes more flat as $\beta$ approaches zero which corresponds to a less informative prior. According to the assumed prior distribution, $\sum_{i=1}^n \theta_i \sim \mathrm{Gamma}(n\alpha,\beta)$ and, hence, the mode of $\sum_{i=1}^n \theta_i$ is $(n\alpha-1)/\beta$. We can interpret the mode of $\sum_{i=1}^n \theta_i $ as the \emph{scale} of the parameter vector.} 

The log-a posteriori probability function can be written as 
\begin{equation}
\rho(\pa) = \ell(\pa) + \ell_0(\pa)
\label{equ:optmap}
\end{equation} 
where $\ell$ is the log-likelihood function in (\ref{equ:btloglik}) and $\ell_0$ is the log-likelihood of the prior distribution given by
\begin{equation}
\ell_0(\pa) = \sum_{i=1}^n\left((\alpha -1) \pae_i - \beta e^{\pae_i}\right).
\label{equ:priorl0}
\end{equation}

Note that for $\alpha = 1$ and $\beta = 0$, the log-a posteriori probability function corresponds to the log-likelihood function. For these values of parameters $\alpha$ and $\beta$, MAP and ML estimation problems are equivalent.  

\paragraph{\bf MM algorithms} The MM algorithm for minimizing a function $f$ is defined by minimizing a surrogate function that \emph{majorizes} $f$. 

A surrogate function $g(\vec{x};\vec{y})$ is said to be \emph{a majorant function} of $f$ if $f(\vec{x}) \leq g(\vec{x};\vec{y})$ and $f(\vec{x}) = g(\vec{x};\vec{x})$ for all $\vec{x}$ and $\vec{y}$. The MM algorithm is defined by the iterative updates: 
\begin{equation}
\vec{x}^{(t+1)} = \arg\min_{\vec{x}} g(\vec{x}; \vec{x}^{(t)}).
\label{equ:mm}
\end{equation}
For maximizing a function $f$, we can analogously define the MM algorithm as minimization of a surrogate function $g$ that \emph{minorizes} function $f$. Majorization surrogate functions are used for minimization of convex functions, and minorization surrogate functions are used for maximization of concave functions.

The classic MM algorithm for the Bradley-Terry model of paired comparisons uses the following minorization function of $\ell(\vec{x})$:
\begin{align}
\underline{\ell}(\vec{x}; \vec{y}) 
=& \sum_{i=1}^n \sum_{j\neq i} \underline{\ell}_{ij}(\vec{x}; \vec{y}) ,
\label{equ:btmin}
\end{align}
where 
$$
\underline{\ell}_{ij}(\vec{x}; \vec{y}) = \win_{i,j}\left(x_i - \frac{e^{x_i} + e^{x_j}}{e^{y_i} + e^{y_j}} - \log\left(e^{y_i} + e^{y_j}\right) + 1\right).
$$
It is easy to observe that $\underline{\ell}(\vec{x}; \vec{y}) $ is a minorization surrogate function of $\ell(\vec{x})$ by noting  that $\log(x) \le x-1$ and that equality holds if, and only if, $x=1$, which is used to break $\log(e^{x_i}+e^{x_j})$ terms in the log-likelihood function.

The \emph{classic MM algorithm} for the ML parameter estimation of the Bradley-Terry model of paired comparisons (\cite{Ford57,hunter04}), is defined by the following iterative updates, for $i = 1,2,\ldots,n$,
\begin{equation}
\theta^{(t+1)}_i = \frac{\sum_{j=1}^n \win_{i,j}}{\sum_{j=1}^n \frac{m_{i,j}}{\theta^{(t)}_i + \theta^{(t)}_j}}.
\label{equ:MMpair}
\end{equation}

Following \cite{Caron12}, the MM algorithm for the MAP parameter estimation of the Bradley-Terry model of paired comparisons is derived for the minorant surrogate function $\underline{\rho}$ of function $\rho$ in (\ref{equ:optmap}), defined as
$$
\underline{\rho}(\vec{x};\vec{y}) =  \underline{\ell}(\vec{x};\vec{y}) + \ell_0(\vec{x})
$$
where $\underline{\ell}(\vec{x};\vec{y})$ is the minorant surrogate function of the log-likelihood function (\ref{equ:btmin}) and $\ell_0$ is the prior log-likelihood function (\ref{equ:priorl0}). 

The iterative updates of the MM algorithm are defined by, for $i=1,2,\ldots,n$, 
\begin{equation}
\theta^{(t+1)} = \frac{\alpha -1 + \sum_{j\neq i} \win_{i,j}}{\beta +\sum_{j\neq i}\frac{m_{i,j}}{\theta_i^{(t)} + \theta_j^{(t)}}}. 
\label{equ:classM}
\end{equation}

Note that this iterative optimization algorithm corresponds to the classic MM algorithm for ML estimation (\ref{equ:MMpair}) when $\alpha = 1$ and $\beta = 0$.

We also consider \emph{gradient descent algorithm} with constant step size $\eta > 0$, which has iterative updates as 
\begin{equation}
\vec{x}^{(t+1)} = \vec{x}^{(t)} - \eta \nabla f(\vec{x}^{(t)}).
\label{equ:gd}
\end{equation}

Our goal in this paper is to characterize the rate of convergence of MM algorithms for generalized Bradley-Terry models. We also derive convergence rates for gradient descent algorithms. It is natural to consider gradient descent algorithms as they belong to the class of first-order optimization methods (not requiring second-order quantities such as Hessian of the objective function or its approximations). Intuitively, the rate of convergence of an iterative algorithm quantifies how fast the value of the objective function converges to the optimum value with the number of iterations. 

For an iterative optimization method for minimizing function $f$, which outputs a sequence of points $\vec{x}^{(0)}, \vec{x}^{(1)}, \ldots $, we say that there is \emph{an $\alpha$-improvement with respect to $f$ at time step $t$} if 
$$
f(\vec{x}^{(t+1)}) - f(\vec{x}^\star) \leq (1-\alpha) (f(\vec{x}^{(t)})-f(\vec{x}^\star))
$$
where $\vec{x}^\star$ is a minimizer of $f$. An iterative optimization method is said to have \emph{linear convergence} if there exist positive constants $\alpha$ and $t_0$ such that the method yields an $\alpha$-improvement at each time step $t\geq t_0$. 

\paragraph{\bf Background on convex analysis} We define some basic concepts from convex analysis that we will use throughout the paper. 

Function $f: \reals^n \rightarrow \reals$ is \emph{$\gamma$-strongly convex} on $\mathcal{X}$ if it satisfies the following subgradient inequality, for all $\vec{x}, \vec{y}\in \mathcal{X}$:
\begin{equation*}
f(\vec{x})-f(\vec{y} ) \leq \nabla f(\vec{x})^{\top} (\vec{x}-\vec{y}) -\frac{\gamma}{2}||\vec{x}-\vec{y}||^2.
\end{equation*}
$f$ is $\gamma$-strongly convex on $\mathcal{X}$ if, and only if, $f(\vec{x})-\frac{\gamma}{2}||\vec{x}||^2$ is convex on $\mathcal{X}$. 

Function $f$ is \emph{$\mu$-smooth} on $\mathcal{X}$ if its gradient $\nabla f$ is $\mu$-Lipschitz on $\mathcal{X}$, i.e., for all $\vec{x},\vec{y}\in \mathcal{X}$,
$$
||\nabla f(\vec{x}) - \nabla f(\vec{y})|| \leq \mu ||\vec{x}-\vec{y}||.
$$
For any $\mu$-smooth function $f$ on $\mathcal{X}$, we also have that, for all $\vec{x}, \vec{y}\in \mathcal{X}$, 
\begin{equation}
|f(\vec{x})-f(\vec{y})-\nabla f(\vec{y})^\top (\vec{x}-\vec{y})| \le  \frac{\mu}{2}||\vec{x}-\vec{y}||^2.
\label{equ:bub}
\end{equation}
A proof of the last claim can be found in Lemma~3.4 in \cite{Bubeck15}.

Function $f$ satisfies \emph{the Polyak-Lojasiewicz inequality} on $\mathcal{X}$ (\cite{P63}) if there exists $\gamma > 0$ such that for all $\vec{x}\in \mathcal{X}$,
\begin{equation}
f(\vec{x})-f(\vec{x}^*) \leq \frac{1}{2\gamma}||\nabla f(\vec{x})||^2 
\label{equ:PL}
\end{equation}
where $\vec{x}^*$ is a minimizer of $f$. When the PL inequality holds on $\mathcal{X}$ for a specific value of $\gamma$, we say that \emph{$\gamma$-PL inequality} holds on $\mathcal{X}$. If $f$ is $\gamma$-strongly convex on $\mathcal{X}$, then $f$ satisfies the $\gamma$-PL inequality on $\mathcal{X}$.

\section{Convergence rates} 
\label{sec:conv-bt} 

In this section, we present results on the rate of convergence for gradient descent and MM algorithms for ML and MAP estimation for the Bradley-Terry model of paired comparisons. We first show some general convergence theorems that hold for any strongly convex and smooth function $f$, which characterize the rate of convergence in terms of the strong-convexity and smoothness parameters of $f$, and a parameter of the surrogate function used to define the MM algorithm. These results are then used to derive convergence rate bounds for the Bradley-Terry model.

The results of this section can be extended to other instances of generalized Bradley-Terry models, including the Rao-Kupper model of paired comparisons with tie outcomes, the Luce choice model, and the Plackett-Luce ranking model. These extensions are established by following the same main steps as for the Bradley-Terry model of paired comparisons. The differences lie in the characterization of the strong-convexity and smoothness parameters. The results provide characterizations of the convergence rates that are equivalent to those for the Bradley-Terry model of paired comparisons up to constant factors. We provide details in Section~\ref{sec:genBT-app}.

\subsection{General convergence theorems}
\label{sec:conv-bt-gen}

We first present a known result that for any $\mu$-smooth function satisfying the $\gamma$-PL inequality, gradient descent algorithm (\ref{equ:gd}) with suitable choice of the step size $\eta$ has a linear convergence with the rate of convergence $1-\gamma/\mu < 1$. This result is due to \cite{N12} and a simple proof can be found in \cite{BV04}, Chapter 9.3. This result follows from the following theorem.

\begin{thm}[gradient descent] Assume $f$ is a convex $\mu$-smooth function on $\mathcal{X}_\mu$ satisfying the $\gamma$-PL inequality on ${\mathcal X}_\gamma\subseteq {\mathcal X}_\mu$, $\vec{x}^*\in {\mathcal X}_\gamma$ is a minimizer of $f$, and $\vec{x}^{(t)}\mapsto \vec{x}^{(t+1)}$ is according to the gradient descent algorithm (\ref{equ:gd}) with step size $\eta = 1/\mu$. 

Then, if $\vec{x}^{(t)}\in \mathcal{X}_\gamma$ and $\vec{x}^{(t+1)}\in {\mathcal X}_\mu$, there is an $\gamma/\mu$-improvement  with respect to $f$ at time step $t$.
\label{thm:conv-gd}
\end{thm}


Note that if there exists $t_0 \geq 0$ such that $\vec{x}^{(t)} \in {\mathcal X}_\gamma$ for all $t\geq t_0$, then Theorem~\ref{thm:conv-gd} implies a linear convergence rate with rate $\gamma/\mu$. Such a $t_0$ indeed exists as it can be shown that $||\vec{x}^{(t)}-\vec{x}^*||$ is non-increasing in $t$ and is decreasing for every $t$ such that $||\nabla f(\vec{x}^{(t)})|| \neq 0$.

We next present a new result which shows that the MM algorithm also has linear convergence, for any smooth and strongly convex function $f$ that has a surrogate function $g$ satisfying a certain condition. 

\begin{thm}[MM] Assume $f$ is a convex $\mu$-smooth function on $\mathcal{X}_\mu$ satisfying the $\gamma$-PL inequality on $\mathcal{X}_\gamma \subseteq \mathcal{X}_\mu$, $\vec{x}^*\in {\mathcal X}_\gamma$ is a minimizer of $f$ and $\vec{x}^{(t)}\mapsto \vec{x}^{(t+1)}$ is according to the MM algorithm (\ref{equ:mm}). Let $g$ be a majorant surrogate function of $f$ such that for some $\delta > 0$, 
$$
g(\vec{x}; \vec{y}) - f(\vec{x}) \leq \frac{\delta}{2} ||\vec{x}-\vec{y}||^2 
\hbox{ for all } \vec{x},\vec{y}\in \mathcal{X}_\mu.
$$
Then, if $\vec{x}^{(t)}\in \mathcal{X}_\gamma$ and $\vec{x}^{(t)} - \frac{1}{\mu + \delta} \nabla f(\vec{x}^{(t)})\in \mathcal{X}_\mu$, there is a $\gamma/(\mu+\delta)$-improvement with respect to $f$ at time step $t$.
\label{thm:conv-mm}
\end{thm}


The proof of Theorem~\ref{thm:conv-mm} is based on separating the gap between the objective function value at an iteration and the optimum function value in two components: (a) one due to using a surrogate function and (b) other due to a virtual gradient descent update. The remaining arguments are similar to those of the proof of Theorem~\ref{thm:conv-gd}. It is worth noting that a different set of sufficient conditions for linear convergence of MM algorithms were found in Proposition~2.7 in \cite{M15}. These conditions require that $g$ is a first-order surrogate function of $f$, a notion we discuss further in Section~\ref{app:mm}. 

From Theorems \ref{thm:conv-gd} and \ref{thm:conv-mm}, we observe that the MM algorithm has the same rate of convergence bound as the gradient descent algorithm except for the smoothness parameter $\mu$ being enlarged for value $\delta$. If $\delta \leq c\mu$, for a constant $c > 0$, then the MM algorithm has the same rate of convergence bound as the gradient descent algorithm up to a constant factor. 

\subsection{Maximum likelihood estimation} 

We consider the rate of convergence for the ML parameter estimation for the Bradley-Terry model of paired comparisons. This estimation problem amounts to finding a parameter vector that minimizes the negative log-likelihood function, with the log-likelihood function given in (\ref{equ:btloglik}). Recall that $\mat{M}$ denotes the matrix of the item-pair co-occurrence counts and $\mat{L}_\mat{M}$ denotes the associated Laplacian matrix. For any positive semidefinite matrix $\mat{A}$, we let $\lambda_i (\mat{A})$ denote the $i$-th smallest eigenvalue of $\mat{A}$.

\begin{lm} For any $\bo \geq 0$, the negative log-likelihood function for the Bradley-Terry model of paired comparisons is $\gamma$-strongly convex on $\mathcal{W}_{\bo,0} = \mathcal{W}_\bo \cap \{\pa\in \reals^n : \pa^{\top} \vec{1} = 0\}$, where $\mathcal{W}_{\bo} = \{\pa\in \reals^n : ||\pa||_\infty\leq \bo\}$, and $\mu$-smooth on $\reals^n$ with
$$
\gamma = c_\bo \lambda_2(\mat{L}_{\mat{M}})  \hbox{ and } \mu = \frac{1}{4}\lambda_n(\mat{L}_{\mat{M}})
$$
where $c_\bo= 1/(e^{-\bo}+e^\bo)^2$.
\label{lm:gammamu}
\end{lm}


By Lemma~\ref{lm:gammamu}, the smoothness parameter $\mu$ is proportional to the largest eigenvalue of the Laplacian matrix $\vec{L}_{\vec{M}}$. By the Gershgorin circle theorem, e.g., Theorem 7.2.1 in \cite{golub}, we have $\lambda_n(\vec{L}_{\vec{M}})\leq 2\dM$. Thus, we can take $\mu = \dM/2$. We will express all our convergence time results in terms of $\dM$ instead of $\lambda_n(\vec{L}_{\vec{M}})$. This is a tight characterization up to constant factors. When $\vec{M}$ is a graph adjacency matrix, then $\lambda_n(\vec{L}_{\vec{M}})\geq \dM + 1$ by \cite{GMS90}. In the context of paired comparisons, $\dM$ has an intuitive interpretation as the maximum number of observed paired comparisons involving an item.

The following lemma will be useful for showing that a function $f$ satisfies the $\gamma$-PL inequality if it satisfies a $\gamma$-strong convexity condition. 

\begin{lm} Assume that $\mathcal{X}$ is a convex set such that $f$ is $\gamma$-strongly convex on $\mathcal{X}_0 = \mathcal{X}\cap \{\vec{x}\in \reals^n : \vec{x}^\top \vec{1} = 0\}$, and that for all $c\in \reals$ and $\vec{x}\in\mathcal{X}$, 
\begin{description}
\item[(C1)] $f(\Pi_c(\vec{x})) = f(\vec{x})$ and
\item[(C2)] $\nabla f(\Pi_c(\vec{x})) = \nabla f(\vec{x})$
\end{description}
where
$$
\Pi_c(\vec{x}) = \vec{x} + c\vec{1}.
$$

Then, $f$ satisfies the $\gamma$-PL inequality on $\mathcal{X}$.
\label{lem:plsuff}
\end{lm}

The proof of Lemma~\ref{lem:plsuff} follows by noting that if $f$ is $\gamma$-strongly convex on $\mathcal{X}_0$, then it satisfies the $\gamma$-PL inequality on $\mathcal{X}_0$. Since for every $\vec{x}\in \mathcal{X}$, $\vec{x} = \vec{x}' + c\vec{1}$ for some $\vec{x}'\in \mathcal{X}_0$ and $c\in \reals$, by conditions (C1) and (C2) and definition of $\gamma$-PL inequality in (\ref{equ:PL}), it follows that if $\gamma$-PL inequality holds on $\mathcal{X}_0$, it holds as well on $\mathcal{X}$. Conditions (C1) and (C2) are satisfied by negative log-likelihood functions for generalized Bradley-Terry models.

Since the negative log-likelihood function of the Bradley-Terry model of paired comparisons  satisfies conditions (C1) and (C2) of Lemma~\ref{lem:plsuff}, combining with Lemma~\ref{lm:gammamu}, we observe that it satisfies the $\gamma$-PL inequality on $\mathcal{W}_\bo$ with $\gamma = c_\bo \aM$. Furthermore, by Lemma~\ref{lm:gammamu}, the negative log-likelihood function is $\mu$-smooth on $\reals^n$ with $\mu = \dM/2$. Combining these facts with Theorem~\ref{thm:conv-gd}, we have the following corollary:

\begin{cor}[gradient descent] Assume that $\pa^*$ is the maximum likelihood parameter estimate in $\mathcal{W}_\bo = \{\pa: \reals^n: ||\pa||_\infty \leq \bo\}$, for some $\bo \geq 0$, and $\pa^{(t)}\mapsto \pa^{(t+1)}$ is according to gradient descent algorithm with step size $\eta = 2/\dM$. 

Then, if $\pa^{(t)}\in \mathcal{W}_\bo$, there is an $\alpha_{\vec{M},\bo}$-improvement at time step $t$ where
$$
\alpha_{\vec{M},\bo} = 2c_\bo\frac{\aM}{\dM}.
$$
\label{cor:gd}
\end{cor}

The result in Corollary~\ref{cor:gd} implies a linear convergence with the rate of convergence bound $1-2c_\bo \aM/\dM$. Hence, we have the following convergence time bound: 
\begin{equation}
T = O\left(\frac{d(\vec{M})}{a(\vec{M})} \log\left(\frac{1}{\epsilon}\right)\right).
\label{equ:convtime}
\end{equation}

We next consider the classic MM algorithm for the ML estimation problem, which uses the surrogate function in (\ref{equ:btmin}). This surrogate function satisfies the following property:

\begin{lm} For any $\bo \geq 0$, for all $\vec{x}, \vec{y} \in [-\bo,\bo]^n$, $\underline{\ell}(\vec{x};\vec{y}) - \ell (\vec{x}) \ge -\frac{\delta}{2} \| \vec{x} - \vec{y} \|^2$ where
$$
\delta = \frac{1}{2} e^{2\bo} \dM.
$$ 
\label{lm:residual}
\end{lm}



By Theorem~\ref{thm:conv-mm} and Lemmas  \ref{lm:gammamu}, \ref{lem:plsuff}, and \ref{lm:residual}, we have the following corollary:

\begin{cor}[MM] Assume that $\pa^*$ is the maximum likelihood parameter estimate in $\mathcal{W}_\bo = \{\pa: \reals^n: ||\pa||_\infty \leq \bo\}$, for some $\bo \geq 0$, and that $\pa^{(t)}\mapsto \pa^{(t+1)}$ is according to the MM algorithm. 

Then, if $\pa^{(t)}\in \mathcal{W}_\bo$, there is an $\alpha_{\vec{M},\bo}$-improvement with respect at time step $t$ where
$$
\alpha_{\vec{M},\bo} = 2c'_\bo\frac{\aM}{\dM}
$$
and
$c'_\bo = 1/[(e^{-\bo}+e^\bo)^2 (1+e^{2\bo})]$.
\label{cor:mm}
\end{cor}

From Corollaries \ref{cor:gd} and \ref{cor:mm}, we observe that both gradient descent and MM algorithms have the rate of convergence bound of the form $1-c \aM/\dM$ for some constant $c > 0$. The only difference is the value of constant $c$. Hence, both gradient descent and MM algorithm have a linear convergence, and the convergence time bound (\ref{equ:convtime}).

\subsection{Maximum a posteriori probability estimation} We next consider the maximum a posteriori probability estimation problem. We first note that the negative log-a posteriori probability function has the following properties. 

\begin{lm}
The negative log-a posteriori probability function for the Bradley-Terry model of paired comparisons and the prior distribution Gamma$(\alpha,\beta)$  is $\gamma$-strongly convex and $\mu$-smooth on $\mathcal{W}_\bo = \{\pa\in\reals^n: ||\pa||_\infty \leq \bo\}$ with
$$
\gamma = e^{-\bo}\beta \hbox{ and } \mu = \frac{1}{4}\lambda_n(\mat{L}_{\mat{M}}) + e^{\bo}\beta.
$$
\label{lm:mapparam}
\end{lm}


Note that the strong convexity parameter $\gamma$ is proportional to $\beta$ while as shown in Lemma~ \ref{lm:gammamu}, for the ML objective $\gamma$ is proportional to $\lambda_2(\mat{M})$. This has important implications on the rate of convergence which we discuss next. 

By Theorem~\ref{thm:conv-gd} and Lemma~\ref{lm:mapparam}, we have the following corollary:

\begin{cor}[gradient descent] Assume $\pa^*$ is the maximum a posteriori parameter estimate in $\mathcal{W}_\bo = \{\pa\in\reals^n: ||\pa||_\infty \leq \bo\}$, for some $\bo \geq 0$, and $\pa^{(t)}\mapsto \pa^{(t+1)}$ is according to gradient descent algorithm (\ref{equ:gd}) with step size $\eta = 2/(\dM + 2\beta e^\bo)$. 

Then, if $\pa^{(t)} \in \mathcal{W}_\bo$, there is a $\alpha_{\vec{M},\bo}$-improvement where
$$
\alpha_{\vec{M},\bo}=\frac{2 e^{-\bo}\beta}{\dM + 2e^\bo\beta}.
$$
\label{cor:gdmap}
\end{cor}

The result in Corollary~\ref{cor:gdmap} implies a linear convergence with the convergence time bound
\begin{equation}
T = O\left(\left(1+\frac{\dM}{\beta}\right)\log\left(\frac{1}{\epsilon}\right)\right).
\label{equ:mapconvtime}
\end{equation}
This bound can be arbitrarily large by taking parameter $\beta$ to be small enough. In Section~\ref{sec:tight}, we will show a simple instance for which this bound is tight. Hence, the convergence time for MAP estimation can be arbitrarily slow, and much slower than for the ML case. 

We next consider the MM algorithm. First, note that since $\underline{\rho}(\vec{x};\vec{y})-\rho(\vec{x}) = \underline{\ell}(\vec{x};\vec{y}) - \ell(\vec{x})$, by Lemma~\ref{lm:residual}, we have 

\begin{lm} For all $\vec{x}, \vec{y} \in [-\bo,\bo]^n$, $\underline{\rho}(\vec{x};\vec{y}) - \rho (\vec{x}) \ge -\frac{\delta}{2} \| \vec{x} - \vec{y} \|^2$ where
$
\delta = \frac{1}{2} e^{2\bo} \dM.
$ 
\label{lm:residual2}
\end{lm}

By Theorem~\ref{thm:conv-mm} and Lemmas \ref{lm:mapparam} and \ref{lm:residual2}, we have the following corollary:

\begin{cor}[MM] Assume $\pa^*$ is the maximum a posteriori parameter estimate in $\mathcal{W}_\bo = \{\pa \in \reals^n: ||\pa||_\infty \leq \bo\}$, for some $\bo \geq 0$, and $\pa^{(t)}\mapsto \pa^{(t+1)}$ is according to the MM algorithm. 

Then, if $\pa^{(t)} \in \mathcal{W}_\bo$, there is an $\alpha_{\vec{M},\bo}$-improvement at time step $t$ where
$$
\alpha_{\vec{M},\bo,\beta} = \frac{2e^{-\bo}\beta}{(1+e^{2\bo})\dM + 2 e^\bo \beta}.
$$
\label{cor:conv-mm-map}
\end{cor}

From Corollaries \ref{cor:mm} and \ref{cor:conv-mm-map}, we observe that both gradient descent algorithm and MM algorithm have the rate of convergence bound $1 - \Omega(\beta/(\beta+\dM))$, and hence both have linear convergence and both have the convergence time bound (\ref{equ:mapconvtime}). We next establish tightness of this rate of convergence by showing that it is achieved for a simple instance.

\subsection{Tightness of the rate of convergence}
\label{sec:tight}

Consider an instance with two items, which are compared $m$ times. Note that $\dM = m$ and $\aM=2m$. This simple instance allows us to express the parameter vector estimate $\pa^{(t)}$ and the log-a posteriori probability function $\rho(\pa^{(t)})$ in a closed form, with detailed derivations provided in Section \ref{app:tight}. From this analysis, we have that $\rho(\pa^{(t)})$ satisfies
$$
\lim_{t\rightarrow\infty}\frac{\rho(\pa^*)-\rho(\pa^{(t+1)})}{\rho(\pa^*)-\rho(\pa^{(t)})} \\
= \left(1+\frac{2(\alpha-1)}{m}\right)^{-2}.
$$

Note that the rate of convergence is approximately $1 - 4(\alpha-1)/m$ for small $\alpha - 1$. By taking $\alpha$ such that $\alpha - 1 = c\beta$, for a constant $c>0$ such that $||\pa^*||_\infty\leq \bo$, we have the rate of convergence $1-\Theta(\beta/\dM)$. This establishes the tightness of the rate of convergence bound in Corollary~\ref{cor:conv-mm-map}.

\subsection{A simple illustrative numerical example} 

\begin{figure}
\begin{center}
\includegraphics[width=0.4\textwidth]{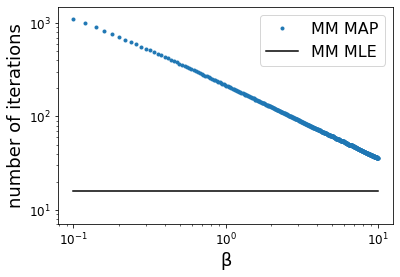}
\end{center}
\caption{Simple illustrative example: number of iterations until convergence versus parameter $\beta$. Note that (a) the smaller the value of $\beta$, the slower the convergence for MAP and (b) the MM algorithm for MAP can be slower for several orders of magnitude than for ML. }
\label{fig:mmconv}
\end{figure}

We illustrate the rate of convergence for a simple example, using randomly generated observations of paired comparisons. This allows us to demonstrate how the number of iterations grows as the value of parameter $\beta$ becomes smaller, and how the number of iterations is affected by the value of parameter $\omega$. Later, in Section~\ref{sec:num}, we provide further validation by using real-world datasets. 

Our example is for an instance with $10$ items with each distinct pair of items compared $10$ times and the input data generated according to the Bradley-Terry model of paired comparisons with the parameter vector such that a half of items have parameter value $-\omega$ and the other half of items have parameter value $\omega$, for a parameter $\omega > 0$. We define the convergence time $T$ to be the smallest integer $t$ such that $||\pa^{(t)}-\pa^{(t-1)}||_\infty \leq \xi$, for a fixed parameter $\xi > 0$. In our experiments, we set $\xi= 0.0001$. 

The results in Figure~\ref{fig:mmconv}, obtained for $\omega = 1/2$, demonstrate that the MM algorithm for the MAP estimation problem with $\beta > 0$ can be much slower than the MM algorithm for the ML estimation problem.

\begin{figure*}[t]
\begin{center}
\includegraphics[width=0.32\textwidth]{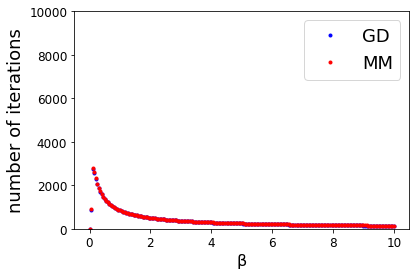}
\includegraphics[width=0.32\textwidth]{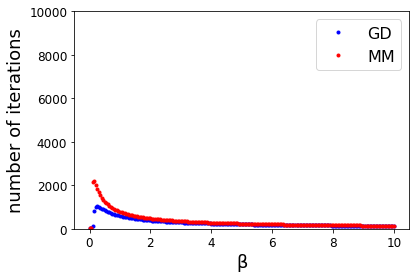}
\includegraphics[width=0.32\textwidth]{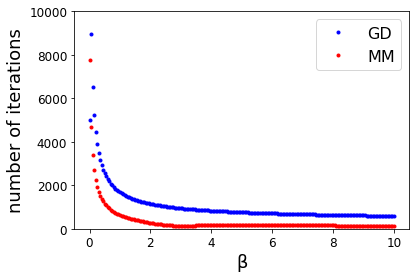}\\
(a) $\omega = 0.01$ \hspace{3.5cm}
(b) $\omega = 1$ \hspace{3.5cm}
(c) $\omega = 5$
\end{center}
\caption{Number of iterations versus $\beta$ for gradient descent and MM algorithms, for different values of $\omega$.}
\label{fig:cgconv}
\end{figure*}

We further evaluate the convergence time of gradient descent and MM algorithms for different values of parameter $\omega$, for each distinct pair of items compared $100$ times. The numerical results in Figure~\ref{fig:cgconv} show the number of iterations versus the value of parameter $\beta$ for gradient descent and MM algorithms, for different values of parameter $\omega$. We observe that for small enough value of $\omega$, convergence times of gradient descent and MM algorithms are nearly identical. Both algorithms have the convergence time increasing by decreasing the value of $\beta$ for strictly positive values of this parameter. We also observe a discontinuity in convergence time, for $\beta = 0$ (MLE case) being smaller than for some small positive value of $\beta$ (MAP case). 

The discontinuity at $\beta = 0$ originates from the fact that the log-likelihood function has infinitely many solutions for $\beta = 0$, but has a unique solution whenever $\beta >0$. Consider a simple illustrative example: $f(x_1,x_2) =  (x_1 - 1)^2 + \theta (x_2 -1)^2$, for a parameter $\theta \geq 0$. Then, the gradient descent converges to the unique solution $(1,1)$ slowly when $\theta$ is close to $0$. When $\theta=0$, however, we just need to find the minimum point of $(x_1-1)^2$ which can be solved in a few iterations.

\section{Accelerated MAP inference} 
\label{sec:acc} 

In this section, we present a new accelerated algorithm for gradient descent and MM algorithms for MAP estimation. \mv{The key element is a transformation of the parameter vector estimate in each iteration of an iterative optimization algorithm that (a) ensures monotonic improvement of the optimization objective along the sequence of parameter vector estimates and (b) ensures certain second-order properties of the objective function hold along the sequence of parameter vector estimates}.
 
We first introduce transformed versions of gradient descent and MM algorithms. Given a mapping $\Pi : \reals^n \rightarrow \reals^n$, we define the \emph{$\Pi$-transformed gradient descent algorithm} by 
\begin{equation}
\vec{x}^{(t+1)} = \Pi(\vec{x}^{(t)} - \eta f(\vec{x}^{(t)})).
\label{equ:accgd}
\end{equation}

Similarly, we define the \emph{$\Pi$-transformed MM algorithm} by the iteration:
\begin{equation}
\vec{x}^{(t+1)} = \Pi(\arg\min_{\vec{x}} g(\vec{x};\vec{x}^{(t)})).
\label{equ:accmm}
\end{equation}

Importantly, function $f$ and mapping $\Pi$ have to satisfy certain conditions in order to provide a convergence rate guarantee, which we discuss in the following section. 

\subsection{General convergence theorems} \label{sec:accgen} Assume that $f$ and $\Pi$ satisfy the following conditions, for a convex set $\mathcal{X}$ that contains optimum point $\vec{x}^*$, and a vector $\vec{d}\in \reals^n$: 
\begin{description}
\item[(F1)] $f$ is $\mu$-smooth on $\mathcal{X}$; 
\item[(F2)] $f$ satisfies the $\gamma$-PL inequality on 
\begin{equation}
\mathcal{X}_0 = \mathcal{X}\cap \{\vec{x}\in \reals^n: \nabla f(\vec{x})^\top \vec{d} = 0\}
\label{equ:x0}
\end{equation}
\end{description}
and
\begin{description}
\item[(P1)] $f(\Pi(\vec{x}))\leq f(\vec{x})$ for all $\vec{x}\in \reals^n$; 
\item[(P2)] $\Pi(\vec{x}) \in \mathcal{X}_0$ when $\Pi(\vec{x}) \in \mathcal{X}$. 
\end{description}
Condition (F1) is a standard smoothness condition imposed on $\mathcal{X}$. Condition (F2) is a standard $\gamma$-PL condition imposed on the subset of points in $\mathcal{X}$ at which the gradient of the function $f$ is orthogonal to vector $\vec{d}$. Condition (P1) means that applying $\Pi$ to a point cannot increase the value of function $f$. \mv{This condition is crucial to ensure a monotonic improvement of the objective function value when transformation $\Pi$ is applied to an iterative optimization method.} Condition (P2) is satisfied when at any $\Pi$-transformed point, the gradient of $f$ is orthogonal to vector $\vec{d}$. \mv{This condition is crucial to ensure certain second-order properties hold when $\Pi$ is applied to an iterative optimization method.}

We have the following two theorems.

\begin{thm}[Gradient descent] Assume that $f$ satisfies (F1) and (F2), $\Pi$ satisfies (P1), and $\eta = 1/\mu$. Let $\vec{x}^{(t)}\mapsto \vec{x}^{(t+1)}$ be according to the $\Pi$-transformed gradient descent algorithm (\ref{equ:accgd}). 

Then, if $\vec{x}^{(t)},\vec{x}^{(t+1)}\in \mathcal{X}_0$, there is an $\gamma/\mu$-improvement with respect to $f$ at time step $t$.
\label{thm:accgd} 
\end{thm}


The theorem establishes the same rate of convergence as for the gradient descent algorithm in Theorem~\ref{thm:conv-gd} but with the strong convexity condition restricted to points at which the gradient of $f$ is orthogonal to vector $\vec{d}$.

\begin{thm}[MM] Assume that $f$ satisfies (F1) and (F2), $\Pi$ satisfies (P1), and $g$ is a majorant surrogate function of $f$ such that $g(\vec{x};\vec{y})-f(\vec{x})\leq \frac{\delta}{2}||\vec{x}-\vec{y}||^2$. Let $\vec{x}^{(t)}\mapsto \vec{x}^{(t+1)}$ be according to the $\Pi$-transformed MM algorithm (\ref{equ:accmm}). 

Then, if $\vec{x}^{(t)},\vec{x}^{(t+1)}\in \mathcal{X}_0$, there is an $\gamma/(\gamma+\delta)$-improvement with respect to $f$ at time step $t$.
\label{thm:accmm}
\end{thm}

The last theorem establishes the same rate of convergence as for classic MM algorithm in Theorem~\ref{thm:conv-mm}, but with a strong convexity condition imposed only at the points at which the gradient of $f$ is orthogonal to vector $\vec{d}$.

We next present a lemma which will be instrumental in showing that the PL condition in (F1) holds for the MAP estimation problem. 

\begin{lm} Assume $f$ is a convex, twice-differentiable function. Let $\mathcal{X}$ be a convex set, $\mathcal{X}_0$ be defined by (\ref{equ:x0}) for a given vector $\vec{d}$, and $\vec{x}^*\in \mathcal{X}_0$ be a minimizer of $f$. 

If for some positive semidefinite $\vec{A}_{\mathcal{X}}$ such that
\begin{description}
\item[(A1)] $\nabla^2 f(\vec{x}) \succeq \vec{A}_{\mathcal{X}}$ for all $\vec{x}\in \mathcal{X}$, and
\item[(A2)] $\vec{u}^\top \vec{A}_{\mathcal{X}} \vec{v} = 0$  for all $\vec{u}, \vec{v}$ such that 
$$
\vec{u} = (\vec{I}-\vec{P}_{\vec{d}})\vec{z} \hbox{ and } \vec{v} = \vec{P}_{\vec{d}}\vec{z}, \hbox{ for } \vec{z}\in \reals^n
$$
where
$$
\vec{P}_{\vec{d}} = \vec{I} - \frac{1}{||\vec{d}||^2}\vec{d}\vec{d}^\top
$$
\end{description}
then, $f$ satisfies the $\gamma$-PL inequality on $\mathcal{X}_0$ for all $\gamma \le \gamma_0$ with 
$$
 \gamma_0 := \min_{\vec{x}\in \reals^n\setminus \{\vec{0}\}: \vec{d}^\top \vec{x} = 0} \frac{\vec{x}^\top \vec{A}_{\mathcal{X}}\vec{x}}{||\vec{x}||^2}.
$$
\label{lem:gamma}
\end{lm}


Note that $\vec{P}_\vec{d}$ is the projection matrix onto the space orthogonal to vector $\vec{d}$. The value of $\gamma_0$ is maximized when vector $ \vec{d}$ is the eigenvector corresponding to
 the smallest eigenvalue of $\vec{A}_{\mathcal{X}}$. In this case, $\gamma_0$ is the second smallest eigenvalue of $\vec{A}_{\mathcal{X}}$.

\subsection{Convergence rate for the BT model} In this section, we apply the framework developed in the previous section to characterize the convergence rate for the MAP parameter estimation of the Bradley-Terry model of paired comparisons. The MAP parameter estimation problem amounts to finding a parameter vector that maximizes the log-a posteriori probability function $\rho$ defined in (\ref{equ:optmap}). Let the transformation $\Pi$ be defined as
\begin{equation}
\Pi(\vec{x}) = \vec{x} + c(\vec{x}) \vec{1}
\label{equ:map}
\end{equation}
where 
\begin{equation}
c(\vec{x}) = \log\left(\frac{\alpha-1}{\beta}n\right) - \log\left(\sum_{i=1}^n e^{x_i}\right).
\label{equ:cx}
\end{equation}

We next show that $f$ and $\Pi$ satisfy conditions (F1), (F2), (P1), and (P2) for the direction vector $\vec{d} = \vec{1}$. This will allow us to apply Theorems \ref{thm:accgd} and \ref{thm:accmm} to characterize the rate of convergence for $\Pi$-transformed gradient descent and MM algorithms. 

We first show that $f$ satisfies conditions (F1) and (F2) for the set $\mathcal{W}_\bo = \{\pa\in \reals^n: ||\pa||_\infty\leq \bo\}$. Condition (F1) holds because, in Lemma~\ref{lm:mapparam}, we have already shown that $f$ is $\mu$-smooth on $\mathcal{W}_\bo$ with $\mu = \dM/2 + e^\bo \beta$. Condition (F2) can be shown to hold by Lemma~\ref{lem:gamma} as follows. Note that we have $\nabla^2 (f(\pa)) \succeq \vec{A}_{\mathcal{W}_\omega}$, for all $\pa \in \mathcal{W}_\bo$, where $\vec{A}_{\mathcal{W}_\omega} = c_\bo \mat{L}_{\mat{M}} + e^{-\omega}\beta \vec{I}$. The assumptions of  Lemma~\ref{lem:gamma} hold: (A1) holds because $\vec{A}_{\mathcal{W}_\omega}$ is a positive semidefinite matrix, and (A2) holds because $\vec{u}^\top \vec{L}_{\vec{M}} \vec{v} = 0$ (which follows from $\vec{L}_{\vec{M}}\vec{1} = \vec{0}$) and $\vec{u}^\top \vec{I} \vec{v} = \vec{u}^\top \vec{v} = 0$ ($\vec{u}$ and $\vec{v}$ are orthogonal). Since $\gamma_0$ is the smallest eigenvalue of $\vec{A}_{\mathcal{X}}$ on the subspace orthogonal to vector $\vec{1}$, we have
$$
\gamma_0 =  c_\omega \lambda_2(\vec{L}_{\vec{M}}) + e^{-\omega}\beta.
$$
Hence, by Lemma~\ref{lem:gamma}, it follows that $f$ satisfies condition (F2) with $\gamma = c_\omega \lambda_2(\vec{L}_{\vec{M}}) + e^{-\omega}\beta$.

We next show that $\Pi$, defined in (\ref{equ:map}), satisfies conditions (P1) and (P2). These two conditions are shown to hold in the following lemma.

\begin{lm} For all $\pa \in \reals^n$,
\begin{equation}
\rho(\Pi(\pa)) \geq \rho(\pa)
\label{lem:ellpi}
\end{equation}
and
\begin{equation}
\nabla \rho(\Pi(\pa))^{\top} \vec{1} = 0.
\label{lem:ellort}
\end{equation}
\label{lem:rhobounds}
\end{lm}


From Theorem~\ref{thm:accgd}, we have the following corollary:

\begin{cor}[Gradient descent] Assume that $\pa^{(t)}\mapsto \pa^{(t+1)}$ is according to the $\Pi$-transformed gradient descent (\ref{equ:accgd}) for the negative log-a posteriori probability function of the Bradley-Terry model of paired comparisons, with product-form prior distribution such that $e^{w_i} \sim$ Gamma ($\alpha,\beta$), $\alpha \geq 1$ and $\beta > 0$, and $\eta = 2/(||\mat{M}||_\infty + 2e^{\bo}\beta)$. 

Then, there is an $\alpha_{\vec{M},\bo,\beta}$-improvement with respect to $\rho$ at time step $t$ where
$$
\alpha_{\vec{M},\bo,\beta} = \frac{2c_\bo \aM + 2 e^{-\bo}\beta}{\dM + 2e^\bo\beta}
$$
and
$c_\bo = 1/(e^{-\bo}+e^{\bo})^2$.
\label{cor:accgd}
\end{cor}

From Theorem~\ref{thm:accmm}, we have the following corollary:

\begin{cor}[MM] Assume that iterates are according to the $\Pi$-transformed MM (\ref{equ:accgd}) for the negative log-a posteriori probability function of the Bradley-Terry model of paired comparisons, with product-form prior distribution such that $e^{w_i} \sim$ Gamma ($\alpha,\beta$), $\alpha \geq 1$ and $\beta > 0$. 

Then, there is an $\alpha_{\vec{M},\bo,\beta}$-improvement with respect to $\rho$ at time step $t$ where
$$
\alpha_{\vec{M},\bo,\beta} = \frac{2c_\bo \aM + 2 e^{-\bo}\beta}{(1+e^{2\bo})\dM + 2e^\bo\beta}
$$
and
$c_\bo = 1/(e^{-\bo}+e^{\bo})^2$.
\label{cor:accmm}
\end{cor}

\proof{Proof} Condition (F1) holds for $-\rho$ because we have already shown that $-\rho$ is $\mu$-smooth with $\mu = \dM/2 + e^\bo \beta$ on $\mathcal{W}_{\bo}$ and $\delta = e^{2\bo} \dM/2$. Condition (F2) holds by Lemma \ref{lem:gamma} with $\gamma = c_\bo\aM + e^{-\bo}\beta$. Conditions (P1) and (P2) hold by (\ref{lem:ellpi}) and (\ref{lem:ellort}) in Lemma~\ref{lem:rhobounds}, respectively. 
\Halmos\endproof

Note that in the limit of small $\beta$, the convergence rate bounds in Corollaries~\ref{cor:accgd} and \ref{cor:accmm} correspond to the bounds for the ML estimation in Corollaries \ref{cor:gd} and \ref{cor:mm}, respectively. From Corollaries~\ref{cor:accgd} and \ref{cor:accmm}, it follows that for accelerated gradient descent and accelerated MM algorithm, the convergence time satisfies
$$
T = O\left(\min\left\{\frac{\dM}{\aM},\frac{\dM}{\beta}\right\}\log\left(\frac{1}{\epsilon}\right)\right).
$$

\begin{algorithm}[t]
\caption{Accelerated MM algorithm}
\label{accMM}
\begin{algorithmic}[1]
\State \textbf{Initialization}: $\epsilon,\theta, \theta^{\mathrm{prev}}$
\While{$||\theta-\theta^{\mathrm{prev}}||_\infty >\epsilon$}
\State $\theta^{\mathrm{prev}} \leftarrow \theta$
\For{$i = 1, 2,\ldots n$}
\State $\theta^{\mathrm{temp}}_{i} = \frac{\alpha -1 + \sum_{j\neq i}\win_{i,j}}{\beta + \sum_{j\neq i}\frac{m_{i,j}}{\theta_i + \theta_j}}$ \Comment{standard MM}
\EndFor
\For{$i = 1, 2,\ldots,n$}
\State $\theta_i =  \frac{\theta_i^{\mathrm{temp}}}{\sum_{j=1}^n \theta_j^{\mathrm{temp}}}\frac{\alpha-1}{\beta}n$ \Comment{rescaling}
\EndFor
\EndWhile
\end{algorithmic}
\end{algorithm}

For the Bradley-Terry model of paired comparisons with parametrization $\theta = (\theta_1, \ldots, \theta_n)^\top$, where $\theta_i = e^{w_i}$ for $i=1,2,\ldots,n$, the transformation $\Pi$ given by (\ref{equ:map}) is equivalent to a \emph{rescaling} as shown in a procedural form in Algorithm~\ref{accMM}. \mv{This algorithm first performs the standard MM update in Eq.~(\ref{equ:classM}), which is followed by rescaling the resulting intermediate parameter vector such that the parameter vector $\theta$ at every iteration satisfies $\sum_{i=1}^n \theta_i = c$ where $c = n(\alpha-1)/\beta$. This can be interpreted as fixing the scale of parameters to a carefully chosen scale that is dependent on the choice of the prior distribution. Note that the scaling factor $c$ cannot be arbitrarily fixed while still preserving good convergence properties. In particular, selecting the scale $c = 1$ can result in undesired convergence properties. We demonstrate this in Section~\ref{sec:num} through numerical examples.} 

\mv{It turns out that the rescaling in Algorithm~\ref{accMM} is roughly of the same order as the random rescaling suggested in \cite{Caron12}.  Therein, the authors suggested using independent identically distributed rescaling factors across different iteration steps with distribution Gamma($n\alpha, \beta$). This ensures the distribution of $\sum_{i=1}^n \theta_i$ to remain invariant across different iterations, equal to Gamma($n\alpha, \beta$). The mode of this rescaling factor is $(n\alpha-1)/\beta$. This bears a similarity with the rescaling in Algorithm~\ref{accMM}, in particular, with respect to the dependence on parameter $\beta$. Our results show that it suffices to use a simple deterministic rescaling factor to ensure linear convergence. Moreover, using a different rescaling than the one used in Algorithm~\ref{accMM} can result in a lack or slow convergence, which is shown by numerical experiments in Section~\ref{sec:num}.} 

\begin{figure}[t]
\begin{center}
\includegraphics[width=0.4\textwidth]{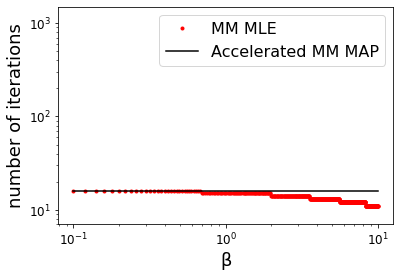}
\end{center}
\caption{The illustrative example revisited: accelerated MM resolves the convergence issue for MAP estimation: it has faster or equal convergence than for ML estimation.}
\label{fig:mmconv2}
\end{figure}

\paragraph{Our numerical example revisited} We ran the accelerated MM algorithm for our numerical example and obtained the results shown in Figure~\ref{fig:mmconv2}. By comparing with the corresponding results obtained by the MM algorithm with no acceleration, shown in Figure~\ref{fig:mmconv}, we observe that the acceleration resolves the slow convergence issue and that it can yield a significant reduction of the convergence time.

\section{Numerical results}
\label{sec:num}

In this section we present evaluation of convergence times of MM algorithms for different generalized Bradley-Terry models for a collection of real-world datasets. Our goal is to provide empirical validation of some of the hypotheses derived from our theoretical analysis. Overall, our numerical results validate that (a) the convergence of the MM algorithm for MAP estimation can be much slower than for ML estimation, (b) MM algorithm for MAP estimation has convergence time that increases as parameter $\beta$ of the prior distribution decreases, and (c) a significant reduction of the convergence time can be achieved by the accelerated MM algorithm defined in Section~\ref{sec:acc}. The code and datasets for reproducing our experiments are available online at:\\

\begin{center}
\url{https://github.com/GDMMBT/AcceleratedBradleyTerry}.
\end{center}

\begin{table}[t]
{\footnotesize
\begin{center}
\caption{Dataset properties.}
\begin{tabular}{|c|c|c|c|c|}
\hline
Dataset & $m$ & $n$ & $d(\vec{M})$ & $a(\vec{M})$  \\ \hline\hline
GIFGIF: A (full) & 161,584 & 6,123 & 83 & 0 \\\hline
GIFGIF: C (full) & 108,126 & 6,122 & 56 & 0  \\\hline
GIFGIF: H (full) & 225,695 & 6,124 &153 & 0 \\\hline\hline
GIFGIF: A (sample) & 702 & 252 & 15 & 0.671  \\\hline
GIFGIF: C (sample) & 734 & 256 & 28 & 0.569  \\\hline
GIFGIF: H (sample) & 1040 &251 & 23 & 1.357 \\ \hline\hline
Chess (full) & 65,030 & 8,631 & 155 & 0 \\\hline
Chess (sample) & 13,181 & 985 & 135 & 1.773 \\ \hline
NASCAR & 64,596 & 83 &1,507 & 39.338 \\ \hline
\end{tabular}
\label{tab:bs}
\end{center}
}
\end{table}

\subsection{Datasets} We consider three datasets, which vary in the type of data, size and sparsity. The three datasets are described as follows.

\paragraph{\bf GIFGIF} This dataset contains user evaluations of digital images by paired comparisons with respect to different metrics, such as amusement, content, and happiness. The dataset was collected through an online web service by the MIT Media Lab as part of the PlacePulse project \cite{gifgif}. This service presents the user with a pair of images and asks to select one that better expresses a given metric, or select none. The dataset contains 1,048,576 observations and covers 17 metrics. We used this dataset to evaluate convergence of MM algorithms for the Bradley-Terry model of paired comparisons. We did this for each of the three aforementioned metrics.

\begin{figure*}[t]
\begin{center}
\includegraphics[width=0.32\textwidth]{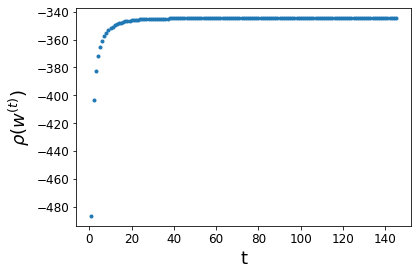} 
\includegraphics[width=0.32\textwidth]{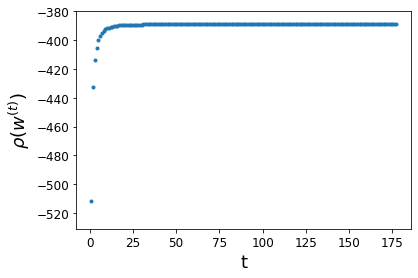} 
\includegraphics[width=0.32\textwidth]{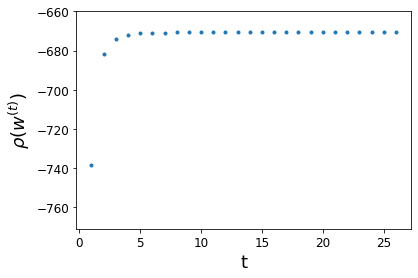}
\includegraphics[width=0.32\textwidth]{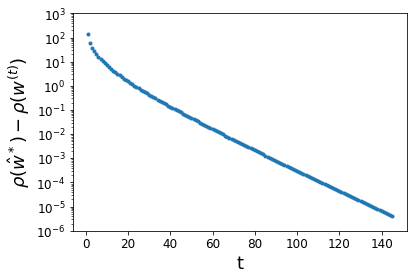} 
\includegraphics[width=0.32\textwidth]{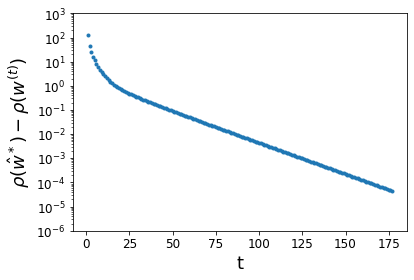} 
\includegraphics[width=0.32\textwidth]{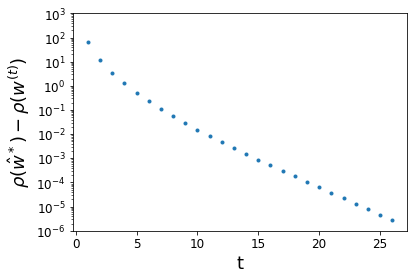}
(a) $\beta = 0$ \hspace{3.5cm}
(b) $\beta = 0.1$ \hspace{3.5cm}
(c) $\beta = 1$\\[5mm]
\caption{\mv{Convergence of the MM algorithm with the input GIFGIF A training dataset for different values of parameter $\beta$: (top) log-a posteriori probability versus the number of iterations, and (bottom) the difference between the maximum log-a posteriori probability and the log a-posteriori probability versus the number of iterations. The maximum a posteriori probability is approximated by taking the parameter vector $\hat{\vec{w}}^*$ output by the MM algorithm after a large number of iterations. The plots in the top row indicate that the algorithm converges. The plots in the bottom row indicate linear convergence. From the plots, we also observe that the convergence is slower for smaller values of parameter $\beta$.}}
\label{fig:linearconvergence}
\end{center}
\end{figure*}

\paragraph{\bf Chess} This dataset contains game-by-game results for 65,030 matches among 8,631 chess players. The dataset was used in a Kaggle chess ratings competition \cite{chess}. Each observation contains information for a match between two players including unique identifiers of the two players, information about which one of the two players played with white figures, and the result of the match, which is either win, loss, or draw. This dataset has a large degree of sparsity. We used this dataset to evaluate convergence of the Rao-Kupper model of paired comparisons with ties. 

\paragraph{\bf NASCAR} This dataset contains auto racing competition results. Each observation is for an auto race, consisting of the ranking of drivers in increasing order of their race finish times. The dataset is available from a web page maintained by \cite{hunter-code}. This dataset was previously used for evaluation of MM algorithms for the Plackett-Luce ranking model by \cite{hunter04} and more recently by \cite{Caron12}. We used this dataset to evaluate convergence times of MM algorithms for the Plackett-Luce ranking model.

We summarise some key statistics for each dataset in Table~\ref{tab:bs}. We use the shorthand notation GIFGIF: A, GIFGIF: C, and GIFGIF: H to denote datasets for metrics amusement, contempt, and happiness, respectively. For full GIFGIF and Chess datasets, we can split the items into two groups such that at least one item in one group is not compared with any item in the other group, i.e., the algebraic connectivity $\aM$ of matrix $\vec{M}$ is zero. \mv{In this case, there exists no ML estimate, while an MAP estimate always exists. In order to consider cases when an MLE exists, we consider sampled datasets by restricting to the set of items such that the algebraic connectivity for this subset of items is strictly positive. This subsampling was done by selecting the largest connected component of items.}

\subsection{Experimental results}

\begin{table*}[t]
\small
\begin{center}
\caption{Number of iterations for the MM algorithm and accelerated MM algorithm (AccMM).}
\begin{tabular}{|c|c|c|c|c|c|c|}
\hline
Dataset & Algorithm & $\beta = 0$ & $0.01$ & $0.1$ & $1$ & $10$ \\ \hline\hline
\multirow{2}{10em}{GIFGIF: A (full)} & MM & MLE & 572 & 125 & 70 &16  \\
& AccMM & non-existant & 509 & 123 & 42 & 13 \\\hline 
\multirow{2}{10em}{GIFGIF: C (full)} & MM & MLE & 733 & 150 & 49 &13  \\ 
& AccMM & non-existant & 551 & 93 & 37 & 13 \\ \hline
\multirow{2}{10em}{GIFGIF: H (full)} & MM & MLE & 1,127 & 149 & 98 & 21  \\
& AccMM & non-existant & 1,044 & 159 & 51 & 18 \\ \hline
\multirow{2}{10em}{GIFGIF: A (sample)} & MM & 145 & 854 & 177 & 26 & 7  \\
& AccMM & & 125  & 81 & 22 & 7 \\ \hline
\multirow{2}{10em}{GIFGIF: C (sample)} & MM & 130 & 694 & 151 & 39 & 9  \\
& AccMM & & 111 & 78 & 36 & 9 \\ \hline
\multirow{2}{10em}{GIFGIF: H (sample)} & MM & 216 & 1,234 & 237 & 38 & 8  \\
& AccMM & & 146 & 72 & 26 & 8 \\ \hline\hline
\multirow{2}{10em}{Chess (full)} & MM & MLE & 2,217 & 581 & 113 & 33  \\
& AccMM & non-existant & 2,291 & 302 & 49 & 25 \\  \hline
\multirow{2}{10em}{Chess (sample)} & MM & 121 & 122 & 91 & 74 & 19  \\
& AccMM & & 117  & 93 & 48 & 16 \\ \hline\hline
\multirow{2}{10em}{NASCAR} & MM & 11 & 695 & 971 & 58 & 10  \\
& AccMM &  & 11 & 11 & 10 & 6 \\  \hline
\multirow{2}{10em}{NASCAR ($\xi=10^{-5}$)} & MM & 14 & 1,528 & 2,069 & 105 & 16  \\
& AccMM &  & 14 & 14 & 12 & {\bf 7} \\  \hline
\multirow{2}{10em}{NASCAR ($\xi=10^{-6}$)} & MM & 17 & 2,362 & 3,223 & 157 & 23  \\
& AccMM &  & 17 & 16 & 14 & 8 \\  \hline
\multirow{2}{10em}{NASCAR ($\xi=10^{-8}$)} & MM & 22 & 4,029 & 5,544 & 261 & 36  \\
& AccMM &  & 22 & 21 & 18 & 11 \\  \hline
\end{tabular}
\label{tab:res}
\end{center}
\end{table*}

We evaluated the convergence time defined as the number of iterations that an algorithm takes until a convergence criteria is satisfied. We use the standard convergence criteria based on the difference of successive parameter vector estimates. \mv{Specifically,  the convergence time $T$ is defined as the smallest integer $t> 0$ such that $||\pa^{(t)} - \pa^{(t-1)}||_\infty \leq \xi$, for fixed value of parameter $\xi > 0$, with initial value $\pa^{(0)} = \vec{0}$. In our experiments, we used $10^{-4}$ as the default value for parameter $\xi$. For NASCAR dataset, we also present results for several other values of $\xi$ to demonstrate how the convergence time changes.} In our experiments, we also evaluated the convergence time measured in real processor time units. We noted that they validate all the observations derived from the convergence times measured in the number of iterations, and hence we do not further discuss them. 

In our experiments we varied the value of parameter $\beta$ and, unless specified otherwise, we set the value of parameter $\alpha$ such that $\alpha -1= \beta$. This corresponds to fixing the mode of the Gamma prior marginal distributions to value $1$. Note that the case $\beta = 0$ corresponds to ML estimation. 

\mv{Before discussing numerical convergence time results, we first show results validating that the MM algorithm converges and that this convergence is linear. This is shown in Figure~\ref{fig:linearconvergence} for GIFGIF A dataset for three different values of parameter $\beta$. For space reasons, we only include results for this dataset. We observe that in all cases the log-a posteriori probability monotonically increases with the number of iterations, thus validating convergence. We also observe that the gap between the maximum log-a posteriori probability and the log-a posteriori probability decreases with the number of iterations in a linear fashion for sufficiently large number of iterations, when plotted using the logarithmic scale for the $y$ axis, thus validating linear convergence.}

\begin{figure*}[t]
\begin{center}
\includegraphics[width=0.32\textwidth]{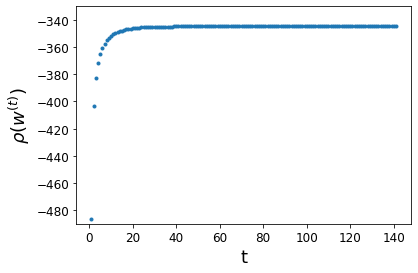} 
\includegraphics[width=0.32\textwidth]{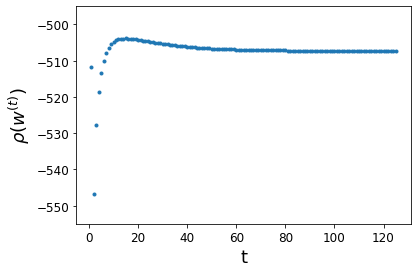} 
\includegraphics[width=0.32\textwidth]{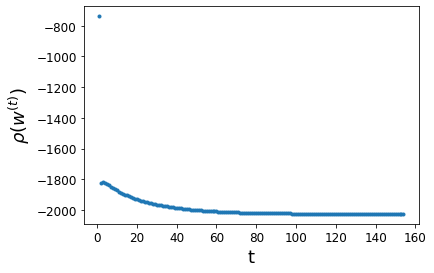}
(a) $\beta = 0$ \hspace{3.5cm}
(b) $\beta = 0.1$ \hspace{3.5cm}
(c) $\beta = 1$\\[5mm]
\end{center}
\caption{\mv{The log-a posteriori probability versus the number of iterations for the MM algorithm with normalization such that $\sum_{i=1}^n e^{w_i} = 1$ at each iteration, using GIFGIF A (sample) as input dataset, for different values of $\beta$. The results indicate that the log-a posteriori probability is not guaranteed to monotonically increase with the number of iterations.}}
\label{fig:mmwithnormalisation}
\end{figure*}

\begin{table*}[t]
\small
\begin{center}
\caption{\mv{Number of iterations for the MM algorithm with normalization such that $\sum_{i=1}^n e^{w_i}=1$ in each iteration.}}
\begin{tabular}{|c|c|c|c|c|c|c|}
\hline
Dataset & Algorithm & $\beta = 0$ & $0.01$ & $0.1$ & $1$ & $10$ \\ \hline\hline
\multirow{1}{10em}{GIFGIF: A (sample)} & MM norm & 141 & 136 & 125 & 154 & No convergence  \\
\hline
\multirow{1}{10em}{GIFGIF: C (sample)} & MM norm & 128 & 130 & 144 & 350 & No convergence  \\
\hline
\multirow{1}{10em}{GIFGIF: H (sample)} & MM norm & 205 & 203 & 184 & 124 & No convergence  \\
\hline\hline

\end{tabular}
\label{tab:mmn}
\end{center}
\end{table*}

We next discuss our numerical results for convergence time evaluated for the MM algorithm and accelerated MM algorithm for different datasets and choice of parameters. Our numerical results are shown in Table~\ref{tab:res}. 


For GIFGIF datasets, we observe that the convergence time increases as the value of parameter $\beta$ decreases for $\beta > 0$. For the values of $\beta$ considered, this increase can be for as much as two orders of magnitude. When the ML estimate exists (for sampled data), we observe that the MM algorithm for ML estimation converges much faster than the MM algorithm for MAP estimation for sufficiently small values of parameter $\beta$. We also observe that a significant reduction of the convergence time can be achieved by the accelerated MM algorithm. This reduction can be for as much as order $10\%$ of the convergence time of the MM algorithm without acceleration. These empirical results validate our theoretical results.

For Chess datasets, all the observations derived by using the GIFGIF datasets remain to hold.

For NASCAR dataset, we show results for different values of parameter $\xi$, including the default value of $10^{-4}$. Again, all the observations made for GIFGIF and Chess datasets remain to hold. It is noteworthy that the MM algorithm for ML estimation converges much faster than for MAP estimation for sufficiently small values of parameter $\beta$. This is especially pronounced for smaller values of $\xi$. For the cases considered, this can be for as much as three orders of magnitude. Similarly, the accelerated MM algorithm converges much faster than the classical MM algorithm.

\mv{In the reminder of this section we highlight the importance of carefully changing the scale of the parameter vector in each iteration, as done in our accelerated MM algorithm, Algorithm~\ref{accMM}, as otherwise the monotonic convergence may not be guaranteed or the convergence may be slow. To demonstrate this, we examine the alternative change of scale such that the parameter vector $\pa$ in each iteration satisfies $\sum_{i=1}^n e^{w_i}=1$. We present  the results for the GIFGIF (sample) datasets. From Figure~\ref{fig:mmwithnormalisation}, we observe that the algorithm does not guarantee a monotonic increase of the a posteriori probability with the number of iterations, which is unlike to our accelerated MM algorithm for which this always holds. In Table~\ref{tab:mmn}, we show the same quantities as in Table~\ref{tab:res} but for the MM algorithm with the alternative change of scale under consideration. We observe that our acceleration method can converge much faster, and that there are cases for which the alternative change of scale results in no convergence within a bound on the maximum number of iterations.}

\section{Further discussion}
\label{sec:disc}

We have shown that for generalized Bradley-Terry models, gradient descent and MM algorithms for the ML estimation problem have a linear convergence with the convergence time bound $O(\dM/\aM)$, where $\dM$ is the maximum number of observed comparisons per item and $\aM$ is the algebraic connectivity of the matrix $\mat{M}$ of the observed counts of item-pair co-occurrences. We have also shown that for generalized Bradley-Terry models, gradient descent and MM algorithms for the MAP estimation problem, with the prior product-form distribution with Gamma$(\alpha,\beta)$ marginal distributions, the convergence time is also linear but with the convergence time bound $O(\dM/\beta)$. This bound is shown to be tight. Our results identify a slow convergence issue for gradient descent and MM algorithms for the MAP estimation problem, which occurs for small values of parameter $\beta$. The small values of parameter $\beta$ correspond to more vague prior distributions. Our results identify a discontinuity of the convergence time at $(\alpha,\beta) = (1,0)$, which corresponds to ML estimation. The proposed acceleration method for the MAP estimation problem resolves the slow convergence issue, and yields a convergence time that is bounded by the best of what can be achieved for the ML and MAP estimation problems. 

\begin{table}[t]
\begin{center}
\caption{Parameters $\dM$ and $\aM$ for matrix $\vec{M}$ being the adjacency matrix of graph $G$ with $n$ vertices}
\label{sphericcase}
{\footnotesize
\begin{tabular}{|c||c|c|c|}
\hline
$G$ & $\dM$ & $\aM$ & $\frac{\dM}{\aM}$ \\
\hline\hline
complete & $n-1$ & $n$ & $\Theta(1)$ \\ \hline
star & $n-1$ & $1$ & $\Theta(n)$ \\ \hline
circuit & $2$ & $2\left(1-\cos\left(\frac{\pi}{n}\right)\right) \sim \frac{4\pi^2}{n^2}$ & $\Theta(n^2)$ \\ \hline
path & $2$ & $2\left(1-\cos\left(\frac{2\pi}{n}\right)\right) \sim \frac{\pi^2}{n^2}$ & $\Theta(n^2)$ 
\\ \hline
\end{tabular}
\label{tab:lambda}
}
\end{center}
\end{table}

Our results provide insights into how the observed comparison data affect the rate of convergence of gradient descent and MM algorithms. The two key parameters affecting the rate of convergence are $\dM$ and $\aM$. For illustration purposes, in Table~\ref{tab:lambda} we show values of $\dM$ and $\aM$ for examples of matrix $\vec{M}$ with $0$-$1$ valued entries, which correspond to graph adjacency matrices. We observe that when each distinct pair is compared the same number of times, i.e. for the complete graph case, the convergence time is $T = O(\log(1/\epsilon))$. For other cases, the convergence time is $T = O(n^c\log(1/\epsilon))$, for some $c\geq 1$. 

\mv{We further consider the case of random design matrices where each distinct pair of items is either compared once or not compared at all, and this is according to independent Bernoulli random variables with parameter $p$ across all distinct pairs of items. In other words, the item pair co-occurrence is according to the Erd\" os-R\' enyi random graph $G_{n,p}$ and $\vec{M}$ is its adjacency matrix. $\dM$ corresponds to the maximum degree of $G_{n,p}$ which has been extensively studied, with precise results obtained for different scalings of $p$ with $n$. In particular, by \cite{BB} (Corollary 3.4), $\dM = pn + O(\sqrt{p n\log(n)})$ with probability $1 - 1/n$ provided that $p = \omega(\log(n)^3/n)$. The algebraic connectivity for Erd\" os-R\' enyi graphs has been studied as well. By \cite{CO07} (Theorem 1.3), $\aM = pn + O(\sqrt{pn\log(n)})$ with probability $1-o(1)$, provided that $p = \omega(\log(n)^2/n)$. Intuitively, if the expected degree $np$ is large enough, $\dM/\aM = \Theta(1)$.}

We can derive an upper bound for the convergence time, which depends only on some simple properties of the graph associated with matrix $\vec{M}$. Let $\vec{A}$ be the adjacency matrix of a graph $G$ which has edge $(i,j)$ if, and only if, $m_{i,j} > 0$. Let $r = \overline{m}/\underline{m}$ where $\overline{m} = \max_{i,j} m_{i,j}$ and $\underline{m} = \min\{m_{i,j}: m_{i,j} > 0\}$. Let $d(n)$ be the maximum degree and $D(n)$ be the diameter of $G$. Then, for both gradient descent and MM algorithms for the ML estimation, we have the convergence time bound (shown in Section~\ref{app:unibound}):
\begin{equation}
T = O(r d(n)D(n)n \log(1/\epsilon)).
\label{equ:unibound}
\end{equation}
This implies that $T = O(r n^3 \log(1/\epsilon))$ for every connected graph $G$, which follows by trivial facts $d(n)\leq n$ and $D(n)\leq n$. The bound in (\ref{equ:unibound}) follows from the lower bound on the algebraic connectivity of a Laplacian matrix $\lambda_2(\vec{L}_{\vec{A}}) \geq 4/(n D(n))$, see Theorem 3.4 in \cite{M94}.


\section{Proofs and Additional Results}
\label{sec:pro}

\subsection{Proof of Theorem~\ref{thm:conv-gd}} \label{app:conv-gd}

Let $\vec{x}'$ be the output of the gradient descent iteration update for input $\vec{x}$ with step size $\eta$. 

If $\vec{x} \in \mathcal{X}_\gamma$ and $\vec{x}'\in {\mathcal X}_\mu$, then
\begin{eqnarray*}
&& f(\vec{x}')-f(\vec{x}^*) \\
& = & f(\vec{x}-\eta \nabla f(\vec{x})) - f(\vec{x}^*)\\
&\leq & f(\vec{x}) - \eta ||\nabla f(\vec{x})||^2 + \frac{\mu}{2}\eta^2 ||\nabla f(\vec{x})||^2 - f(\vec{x}^*)\\
&=& f(\vec{x}) - f(\vec{x}^*) - \left(\eta-\frac{\mu}{2}\eta^2\right)||\nabla f(\vec{x})||^2\\
&\leq & f(\vec{x}) - f(\vec{x}^*) - 2\gamma\left(\eta - \frac{\mu}{2}\eta^2\right)(f(\vec{x})-f(\vec{x}^*))\\
&=& (1-2\gamma \eta + \gamma \mu \eta^2)(f(\vec{x})-f(\vec{x}^*))
\end{eqnarray*}
where the first inequality is by the assumption that $f$ is $\mu$-smooth on $\mathcal{X}_\mu$  and the second inequality is by the assumption that $f$ satisfies the $\gamma$-PL inequality on $\mathcal{X}_\gamma$. Taking $\eta = 1/\mu$, which minimizes the above bound, establishes the claim of the theorem.

\subsection{Proof of Theorem~\ref{thm:conv-mm}}\label{app:conv-mm} 

Let $\vec{x}'$ be the output of the MM algorithm iteration update for input $\vec{x}$.

By the facts $f(\vec{x}')\leq g(\vec{x}'; \vec{x})$ and $g(\vec{x}'; \vec{x})\leq g(\vec{z}; \vec{x})$ for all $\vec{z}$, for any $\eta \geq 0$,
\begin{eqnarray*}
&& f(\vec{x}') - f(\vec{x}^*) \\
&\leq & g(\vec{x}'; \vec{x}) - f(\vec{x}^*)\\
&\leq & g(\vec{x}-\eta \nabla f(\vec{x}); \vec{x}) - f(\vec{x}^*)\\
&=& f(\vec{x} -\eta \nabla f(\vec{x})) - f(\vec{x}^*) \\
&& + g(\vec{x}-\eta \nabla f(\vec{x}); \vec{x}) - f(\vec{x} -\eta \nabla f(\vec{x})).
\end{eqnarray*}


Now, by the same arguments as in the proof of Theorem~\ref{thm:conv-gd}, if $\vec{x}\in \mathcal{X}_\gamma$ and $\vec{x} -\eta \nabla f(\vec{x})\in \mathcal{X}_\mu$, we have
\begin{eqnarray*}
&& f(\vec{x}-\eta \nabla f(\vec{x})) - f(\vec{x}^*) \\
& \leq & (1-2\gamma \eta + \gamma \mu \eta^2)(f(\vec{x})-f(\vec{x}^*)).
\end{eqnarray*}

Next, if $\vec{x}\in \mathcal{X}_\gamma$ and $\vec{x} -\eta \nabla f(\vec{x})\in \mathcal{X}_\mu$,
\begin{eqnarray*}
&& g(\vec{x}-\eta \nabla f(\vec{x}); \vec{x})
-f(\vec{x}-\eta \nabla f(\vec{x})) \\
& \leq & \frac{\delta}{2} \eta^2 ||\nabla f(\vec{x})||^2\\
&\leq & \delta \eta^2 \gamma (f(\vec{x})-f(\vec{x}^*))
\end{eqnarray*}
where the first inequality is by the smoothness condition on the majorant surrogate function and the second inequality is by the assumption that $f$ satisfies the PL inequality with parameter $\gamma$ on $\mathcal{X}_\gamma$.

Putting the pieces together, we have
\begin{eqnarray*}
&& f(\vec{x}') - f(\vec{x}^*) \\
& \leq & \left(1 - 2\gamma \eta + \gamma(\mu+\delta) \eta^2\right) (f(\vec{x})-f(\vec{x}^*)).
\end{eqnarray*}
Taking $\eta = 1/(\mu+\delta)$ (which minimizes the factor involving $\eta$ in the last inequality) yields the asserted result.

\subsection{Comparison of Theorem~\ref{thm:conv-mm} with Proposition~2.7 in \cite{M15}} \label{app:mm}

\begin{thm}[Proposition~2.7 in \cite{M15}] Suppose that $f$ is a strongly convex function on $\mathcal{X}_\gamma$ and $\vec{x}^*$ is a minimizer of $f$ and that it holds $\vec{x}^*\in \mathcal{X}_\gamma$. Assume that $g$ is a first-order surrogate function of $f$ on $\mathcal{X}_\mu$ with parameter $\mu_0 > 0$. Let $\vec{x}^{(t+1)}$ be the output of the MM algorithm for input $\vec{x}^{(t)}$. Then, if $\vec{x}^{(t)}\in \mathcal{X}_\gamma$ and $\vec{x}^{(t+1)}\in \mathcal{X}_\mu$, then we have
$$
f(\vec{x}^{(t+1)}) - f(\vec{x}^*) \leq c (f(\vec{x}^{(t)}) - f(\vec{x}^*))
$$
where
$$
c = \left\{
\begin{array}{ll}
\frac{\mu_0}{\gamma}, & \hbox{ if } \gamma > 2 \mu_0\\
1-\frac{\gamma}{4\mu_0}, & \hbox{ if } \gamma \leq 2\mu_0.
\end{array}
\right .
$$
\label{thm:M15}
\end{thm}

\proof{Proof} If $g$ is a first-order surrogate function on $\mathcal{X}_\mu$ with parameter $\mu_0$, then
$$
f(\vec{x}') \leq f(\vec{z}) + \frac{\mu_0}{2}||\vec{z}-\vec{y}||^2
$$ 
where $\vec{x}' = \arg\min_{\vec{z}'} g(\vec{z}';\vec{y})$.

From this, it follows that 
\begin{eqnarray*}
&& f(\vec{x}') \\
 &\leq & \min_{\vec{z}}\left\{ f(\vec{z}) + \frac{\mu_0}{2}||\vec{z}-\vec{x}^*||^2\right\}\\
& \leq & \min_{a\in[0,1]} \left\{ f(a\vec{x}^* + (1-a)\vec{x}) + \frac{\mu_0 a^2}{2}||\vec{x}-\vec{x}^*||^2\right\}\\
& \leq & \min_{a\in[0,1]} \left\{ a f(\vec{x}^*) + (1-a)f(\vec{x}) + \frac{\mu_0 a^2}{2}||\vec{x}-\vec{x}^*||^2\right\}
\end{eqnarray*}
where the last inequality is by convexity of $f$.

We have established the following inequality
\begin{eqnarray*}
&& f(\vec{x}') - f(\vec{x}^*) \\
& \leq & \min_{a\in [0,1]}\left\{(1-a) (f(\vec{x})-f(\vec{x}^*)) + \frac{\mu_0 a^2}{2}||\vec{x}-\vec{x}^*||^2\right\}.
\end{eqnarray*}

By assumption that $f$ is $\gamma$-strongly convex on $\mathcal{X}_\gamma$ and $\vec{x}\in \mathcal{X}_\gamma$, we have
$$
f(\vec{x})-f(\vec{x}^*) \geq \frac{\gamma}{2}||\vec{x}-\vec{x}^*||^2.
$$

It follows that
\begin{eqnarray*}
&& f(\vec{x}') - f(\vec{x}^*) \\
& \leq & \min_{a\in [0,1]}\left\{1-a + \frac{\mu_0 a^2}{\gamma}\right\} (f(\vec{x})-f(\vec{x}^*)).
\end{eqnarray*}
It remains only to note that
$$
\min_{a\in [0,1]}\left\{1-a + \frac{\mu_0 a^2}{\gamma}\right\} = c.
$$
\Halmos\endproof

The rate of convergence bound derived from Theorem~\ref{thm:conv-mm} can be tighter than the rate of convergence bound derived from Theorem~\ref{thm:M15}. 

To show this consider the Bradley-Terry model for which we have shown in Lemma~\ref{lm:residual} that the surrogate function $\underline{\ell}$ of the log-likelihood function $\ell$ satisfies condition of Theorem~\ref{thm:conv-mm} on $[-\bo,\bo]^n$ with $\delta = \frac{1}{2}e^{2\bo}\dM$. It also holds that surrogate function $\underline{\ell}$ is also a first-order surrogate function of $\ell$ on $[-\bo,\bo]^n$ with $\mu_0 = \frac{1}{2}e^{2\bo}\dM$. Hence in this case, we have $\delta = \mu_0$.

The convergence rate bound of Theorem~\ref{thm:conv-mm} is tighter than the convergence rate bound of Theorem~\ref{thm:M15} if and only if $\mu + \delta < 4\mu_0$. Since $\delta = \mu_0$, this is equivalent to $\mu < 3 \delta$. Since by Lemma~\ref{lm:gammamu} we can take $\mu = \frac{1}{2}\dM$, the latter condition reads as
$$
1 < 3e^\bo
$$
which indeed holds true.

\subsection{Proof of Lemma~\ref{lm:gammamu}}\label{app:gammamu}

The Hessian of the negative log-likelihood function has the following elements:
\begin{equation}
\nabla^2(-\ell(\pa))_{i,j} = \left\{
\begin{array}{ll}
\sum_{v\neq i} m_{i,v}\frac{e^{\pae_i}e^{\pae_v}}{(e^{\pae_i}+e^{\pae_v})^2}, & \hbox{ if } i= j\\
-m_{i,j}\frac{e^{\pae_i}e^{\pae_j}}{(e^{\pae_i}+e^{\pae_j})^2}, & \hbox{ if } i\neq j.
\end{array}
\right .
\label{equ:BThessian}
\end{equation} 
We will show that for all $i\neq j$,
\begin{equation}
\frac{\partial^2}{\partial \pae_i \partial \pae_j}(-\ell(\pa))\leq -c_\bo m_{i,j} \hbox{ for all } \pa \in [-\bo,\bo]^n
\label{equ:nabla2-1}
\end{equation}
and
\begin{equation}
-\frac{1}{4}m_{i,j}\leq \frac{\partial^2}{\partial \pae_i \partial \pae_j}(-\ell(\pa)) \hbox{ for all } \pa \in \reals^n.
\label{equ:nabla2-2}
\end{equation}

From (\ref{equ:nabla2-1}), we have $\nabla^2 (-\ell(\pa)) \succeq c_\bo \mat{L}_{\mat{M}}$ for all $\pa \in [-\bo,\bo]^n$. Hence, for all 
$\pa \in [-\bo,\bo]^n$ and $\vec{x} \in \mathcal{X}$, 
$$
\vec{x}^{\top} \nabla^2 (-\ell(\pa))\vec{x} \geq c_\bo \lambda_2(\mat{L}_{\mat{M}})||\vec{x}||^2 
$$
where $\mathcal{X} = \{\vec{x}\in\reals^n : \vec{x}^{\top} \vec{1} = 0\}$. This shows that $-\ell$ is $c_\bo \lambda_2(\mat{L}_{\mat{M}})$-strongly convex on $\mathcal{X}$.

From (\ref{equ:nabla2-2}), we have $\frac{1}{4} \mat{L}_{\mat{M}}\succeq \nabla^2 (-\ell(\pa))$ for all $\pa\in\reals^n$. Hence, for all $\vec{x}\in \reals^n$,
$$
\vec{x}^{\top} \nabla^2 (-\ell(\pa)) \vec{x} \leq \frac{1}{4}\lambda_n(\mat{L}_{\mat{M}})||\vec{x}||^2.
$$
This shows that $-\ell$ is $\frac{1}{4}\lambda_n(\mat{L}_{\mat{M}})$-smooth on $\reals^n$.

It remains to show that (\ref{equ:nabla2-1}) and (\ref{equ:nabla2-2}) hold. For (\ref{equ:nabla2-1}), we need to show that $c_\bo \leq x_i x_j/(x_i+x_j)^2$ for all $\vec{x}\in [-\bo,\bo]^n$. Note that $x_i x_j/(x_i+x_j)^2 = z(1-z)$ where $z:=x_i /(x_i+x_j)$. Note that $z\in \Omega: = [e^{-\bo}/(e^{-\bo}+e^\bo), 1- e^{-\bo}/(e^{-\bo}+e^\bo)]$ for all $\vec{x}\in [-\bo,\bo]^n$. The function $z(1-z)$ achieves its minimum over the interval $\Omega$ at a boundary of $\Omega$. Thus, it holds $\min_{z\in \Omega} z(1-z) = c_\bo$. For (\ref{equ:nabla2-2}), we can immediately note that for all $\pa \in \reals^n$,
$$
\frac{\pae_i\pae_j}{(\pae_i+\pae_j)^2} = \frac{\pae_i}{\pae_i + \pae_j}\left(1-\frac{\pae_i}{\pae_i + \pae_j}\right) \leq \frac{1}{4}.
$$

\subsection{Proof of Lemma~\ref{lm:residual}}\label{app:residual}

Let $\vec{y}$ be an arbitrary vector in $[-\bo,\bo]^n$. Let $r(\vec{x};\vec{y}) = \underline{\ell}(\vec{x};\vec{y}) - \ell (\vec{x})$ for $\vec{x}\in [-\bo,\bo]^n$. Then, we have 
\begin{equation}
\begin{array}{c}
r(\vec{y};\vec{y}) = 0, \nabla_{\vec{x}} r(\vec{y};\vec{y}) =0, \hbox{ and }\\
 \nabla^2_x r(\vec{x};\vec{y}) = \nabla^2(-\ell(\vec{x})) + A
 \end{array}
\label{equ:ry}
\end{equation}
where $A$ is a $n\times n$ diagonal matrix with diagonal elements 
$$
A_{i,i} = -\sum_{j\in i} m_{i,j} \frac{e^{x_i}}{e^{y_i}+e^{y_j}} \geq -\frac{1}{2}e^{2\bo}||\mat{M}||_\infty. 
$$
Since $\nabla^2 (-\ell(\vec{x}))$ is a positive semi-definite matrix and $A$ is a diagonal matrix, for all $\vec{x},\vec{y}\in [-\bo,\bo]^n$ and $\pa \in [-\bo,\bo]^n$, we have
$$
\vec{x}^{\top} \nabla_{\vec{x}}^2 r(\pa;\vec{y})\vec{x} \geq - ||\mat{M}||_\infty\frac{e^{2\bo}}{2}||\vec{x}||^2 = -\delta ||\vec{x}||^2.
$$
By limited Taylor expansion, for all $\vec{x} \in [-\bo,\bo]^n$,
\begin{eqnarray*}
&& r(\vec{x};\vec{y}) \\
& \geq & r(\vec{y};\vec{y}) + (\vec{x}-\vec{y})^{\top} \nabla_{\vec{x}}r(\vec{y};\vec{y})  \\
&& + \frac{1}{2}\min_{0\leq a \leq 1} (\vec{x}-\vec{y})^{\top} \nabla_{\vec{x}}^2 r(a \vec{x} + (1-a) \vec{y};\vec{y})(\vec{x}-\vec{y})\\
&=& \frac{1}{2}\min_{0\leq a \leq 1} (\vec{x}-\vec{y})^{\top} \nabla_{\vec{x}}^2 r(a \vec{x} + (1-a) \vec{y})(\vec{x}-\vec{y};\vec{y})\\
&\geq & -\frac{\delta}{2}||\vec{x}-\vec{y}||^2.
\end{eqnarray*}
By the definition of $r(\vec{x};\vec{y})$, we have $\bar{\ell}(\vec{x};\vec{y})-\ell(\vec{x})\geq - \frac{\delta}{2}||\vec{x}-\vec{y}||^2$.

\subsection{Surrogate function (\ref{equ:btmin}) for the Bradley-Terry model is a first-order surrogate function} \label{app:surrogate}

We show that the surrogate function $\underline{\ell}$ of the log-likelihood function $\ell$ of the Bradley-Terry model, given by (\ref{equ:btmin}), is a first-order surrogate function on $\mathcal{X}_\bo = [-\bo,\bo]^n$ with $\mu_0 = \frac{1}{2}e^{2\bo}\dM$.

We need to show that the error function $h(\vec{x};\vec{y}) = \ell(\vec{x}) - \underline{\ell}(\vec{x};\vec{y})$ is a $\mu_0$-smooth function on $\mathcal{X}_\bo$.

By a straightforward calculus, we note
$$
\nabla^2 h(\vec{x}; \vec{y}) = \nabla^2 \ell(\vec{x}) + D(\vec{x},\vec{y}) 
$$
where $D(\vec{x},\vec{y})$ is a diagonal matrix with diagonal elements
$$
d_u = \sum_{j\neq u} m_{u,j}\frac{e^{x_u}}{e^{y_u} + e^{y_j}}.
$$
We can take
$$
\mu_0 = \max_{\vec{x},\vec{y}\in \mathcal{X}_\bo} \max\{|\lambda_1(\nabla^2 h(\vec{x}; \vec{y}))|, |\lambda_n(\nabla^2 h(\vec{x}; \vec{y}))|\}.
$$

For any $A = B + D$ where $B$ is a $n\times n$ matrix and $D$ is a $n\times n$ diagonal matrix with diagonal elements $d_1, d_2,\ldots, d_n$, we have
$$
\lambda_1(B) + \min_u d_u \leq \lambda_i(A) \leq \lambda_n(B) + \max_u d_u.
$$
It thus follows that
\begin{eqnarray*}
\mu_0 & \leq & \max_{\vec{x},\vec{y}\in \mathcal{X}_\bo} \max\{|\lambda_1(\nabla^2 \ell(\vec{x}))| \\
&& + \min_u d_u|, |\lambda_n(\nabla^2 \ell (\vec{x})) + \max_{u}d_u|\}.
\end{eqnarray*}

Now note that for all $\vec{x},\vec{y}\in \mathcal{X}_\bo$,
$$
-\frac{1}{2}\dM\leq \lambda_1(\nabla^2 \ell(\vec{x}))\leq \lambda_n(\nabla^2 \ell(\vec{x})) = 0
$$
and
$$
\frac{1}{2}e^{-2\bo}\min_u \sum_{j\in u}m_{u,j} \leq \min_u d_u \leq \max_u d_u \leq \frac{1}{2}e^{2\bo}\dM.
$$

We have
$$
|\lambda_n(\nabla^2 \ell (\vec{x})) + \max_{u}d_u| = \max_u d_u \leq \frac{1}{2}e^{2\bo}\dM
$$
and 
\begin{eqnarray*}
&& |\lambda_1(\nabla^2 \ell (\vec{x})) + \min_{u}d_u| \\
&=& (\lambda_1(\nabla^2 \ell (\vec{x})) + \min_{u}d_u)\ind_{\lambda_1(\nabla^2 \ell (\vec{x})) + \min_{u}d_u \geq 0}\\
&& + (-\lambda_1(\nabla^2 \ell (\vec{x})) - \min_{u}d_u)\ind_{\lambda_1(\nabla^2 \ell (\vec{x})) + \min_{u}d_u < 0}\\
&\leq & \min_u d_u \ind_{\lambda_1(\nabla^2 \ell (\vec{x})) + \min_{u}d_u \geq 0}\\
&& -\lambda_1(\nabla^2 \ell (\vec{x}))\ind_{\lambda_1(\nabla^2 \ell (\vec{x})) + \min_{u}d_u < 0}\\
& \leq & \frac{1}{2}e^{2\bo}\dM \ind_{\lambda_1(\nabla^2 \ell (\vec{x})) + \min_{u}d_u \geq 0}\\
&& + \frac{1}{2}\dM\ind_{\lambda_1(\nabla^2 \ell (\vec{x})) + \min_{u}d_u < 0}\\
&\leq & \frac{1}{2}e^{2\bo}\dM.
\end{eqnarray*}

\subsection{Proof of Lemma~\ref{lm:mapparam}} \label{app:mapparam}

We consider the log-a posteriori probability function $\rho(\pa) = \ell(\pa) + \ell_0(\pa) + \mathrm{const}$ where $\ell$ is the log-likelihood function given by (\ref{equ:btloglik}) and $\ell_0$ is the prior log-likelihood function given by (\ref{equ:priorl0}). Note that $\nabla^2 (-\ell_0(\pa))$ is a diagonal matrix with diagonal elements equal to $\beta e^{\pae_i}$, for $i = 1,2,\ldots,n$. It can be readily shown that for $\pa\in {\mathcal W}_\bo$,
\begin{equation}
c_\bo\mat{L}_{\mat{M}} + e^{-\bo}\beta\mat{I}_n \preceq \nabla^2 (-\rho(\pa))\preceq \frac{1}{4} \mat{L}_{\mat{M}} + e^{\bo}\beta\mat{I}_n. 
\label{equ:bnds}
\end{equation}
From (\ref{equ:bnds}), for all $\pa \in {\mathcal W}_\bo$ and $\vec{x}\in \reals^n$,
$$
\vec{x}^\top \nabla^2(-\rho(\pa))\vec{x} \geq \lambda_1(e^{-\bo}\beta\mat{I}_n)||\vec{x}||^2 = e^{-\bo}\beta ||\vec{x}||^2.
$$
Hence, $-\rho$ is $e^{-\bo}\beta$-strongly convex on ${\mathcal W}_\bo$.

Similarly, from (\ref{equ:bnds}), for all $\pa \in {\mathcal W}_\bo$, and $\vec{x}\in \reals^n$,
\begin{eqnarray*}
\vec{x}^\top \nabla^2(-\rho(\pa))\vec{x} & \leq & \lambda_n( \frac{1}{4} \mat{L}_{\mat{M}} + e^{\bo}\beta\mat{I}_n)||\vec{x}||^2\\
&\leq & (\lambda_n(\frac{1}{4}\mat{L}_{\mat{M}}) + \lambda_n(e^{\bo}\beta\mat{I}_n))||\vec{x}||^2\\
&= & (\frac{1}{4}\lambda_n(\mat{L}_{\mat{M}}) + e^{\bo}\beta)||\vec{x}||^2.
\end{eqnarray*}
Hence, $-\rho$ is $\mu$-smooth on ${\mathcal W}_\bo$ with $\mu = \frac{1}{4}\lambda_n(\mat{L}_{\mat{M}}) + e^{\bo}\beta$.

\subsection{The asymptote in Section \ref{sec:tight}}\label{app:tight}

We consider the case of two items, compared $m$ times. Suppose that the observed data is such that the number of comparisons won by items $1$ and $2$ are $\win_1$ and $\win_2$, respectively. 

The MM algorithm iterates $\pa^{(t)}$ are such that
$$
e^{\pae_i^{(t+1)}} = \frac{\win_i + \alpha - 1}{m + \beta s^{(t)}}s^{(t)}, \hbox{ for } i = 1,2
$$
where $s^{(t)} = e^{\pae_1^{(t)}} + e^{\pae_2^{(t)}}$. Observe that $s^{(t)}$ evolves according to the following autonomous nonlinear dynamical system:
\begin{equation}
s^{(t+1)} = \frac{m + 2(\alpha-1)}{m + \beta s^{(t)}}s^{(t)}.
\label{equ:s}
\end{equation}

The limit point of $s^{(t)}$ as $t$ goes to infinity is $2(\alpha-1)/\beta$. Note that $(\alpha-1)/\beta$ is the mode of Gamma$(\alpha,\beta)$.

Let us define $a^{(t)}$ by $s^{(t)} = [2(\alpha-1)/\beta] (1+a^{(t)})$. Note that $a^{(t)}$ goes to $0$ as $t$ goes to infinity. By a tedious but straightforward calculus, we can show that
\begin{equation}
\rho(\pa^*) - \rho(\pa^{(t)}) = 2(\alpha-1)(a^{(t)} - \log(1+a^{(t)})).
\label{equ:rhodiff}
\end{equation}

From (\ref{equ:s}), note that $1/a^{(t)}$ evolves according to a linear dynamical system, which allows us to derive the solution for $a^{(t)}$ in the explicit form given as follows: 
$$
a^{(t)} = \left(\frac{1}{1-\frac{2(\alpha-1)}{\beta s^{(0)}}}\left(1+\frac{2(\alpha-1)}{m}\right)^t-1\right)^{-1}.
$$
From (\ref{equ:rhodiff}), $\rho(\pa^*)-\rho(\pa^{(t)}) = (\alpha-1)(a^{(t)})^2(1+o(1))$ for large $t$, and thus
\begin{equation*}
\begin{array}{ll}
\rho(\pa^*) - \rho(\pa^{(t)}) 
=& (\alpha-1)\left(1-\frac{2(\alpha-1)}{\beta s^{(0)}}\right)^2\\
& \left(1+\frac{2(\alpha-1)}{m}\right)^{-2t}(1+o(1)).
\end{array}
\end{equation*}
It follows that the rate of convergence of the log-a posteriori probability function is given as follows:
$$
\lim_{t\rightarrow\infty}\frac{\rho(\pa^*)-\rho(\pa^{(t+1)})}{\rho(\pa^*)-\rho(\pa^{(t)})} = \left(1+\frac{2(\alpha-1)}{m}\right)^{-2}.
$$

\subsection{Proof of Theorem~\ref{thm:accgd}}\label{app:accgd}

Since $f(\Pi(\vec{x}))\leq f(\vec{x})$ for all $\vec{x}\in \reals^n$, 
\begin{eqnarray*}
f(\vec{x}^{(t+1)}) &=& f(\Pi(\vec{x}^{(t)} - \eta \nabla f(\vec{x}^{(t)})))\\
& \leq & f(\vec{x}^{(t)} - \eta \nabla f(\vec{x}^{(t)})).
\end{eqnarray*}
By the same steps as those in the proof of Theorem~\ref{thm:conv-gd}, we can show that 
\begin{eqnarray*}
&& f(\vec{x}^{(t)} - \eta \nabla f(\vec{x}^{(t)})) - f(\vec{x}^*)\\
& \leq & \left(1-\frac{\gamma}{\mu}\right) (f(\vec{x}^{(t)}) - f(\vec{x}^*)).
\end{eqnarray*}
Hence, it follows that
\begin{eqnarray*}
&& f(\vec{x}^{(t+1)}) - f(\vec{x}^*) \\
&\leq & \left(1-\frac{\gamma}{\mu}\right)(f(\vec{x}^{(t)}) - f(\vec{x}^*)).
\end{eqnarray*}

\subsection{Proof of Lemma~\ref{lem:gamma}} \label{app:gamma} 

By a limited Taylor expansion, for any $\vec{x},\vec{y}\in \reals^n$, we have
\begin{equation}
\begin{array}{ll}
& f(\vec{y})  \geq  f(\vec{x}) + \nabla f(\vec{x})^{\top}(\vec{y}-\vec{x}) \\
& + \frac{1}{2}\min_{a\in [0,1]} (\vec{y}-\vec{x})^{\top}\nabla^2 f(a \vec{y} + (1-a)\vec{x})(\vec{y}-\vec{x}).
\end{array}
\label{equ:e1}
\end{equation}

Let 
$$
\vec{u} = \left(\vec{I} - \vec{P}_\vec{d}\right) (\vec{y} - \vec{x}) \hbox{ and } \vec{v} = \vec{P}_\vec{d}(\vec{y}-\vec{x})
$$
where
$$
\vec{P}_\vec{d} = \mat{I} -\frac{1}{||\vec{d}||^2}\vec{d}\vec{d}^{\top}.
$$

Notice that 
\begin{enumerate}
\item[(i)] $\vec{u}+\vec{v} = \vec{y} - \vec{x}$, and
\item[(ii)] $\vec{u}$ and $\vec{v}$ are orthogonal, i.e., $\vec{u}^{\top} \vec{v} = 0$.
\end{enumerate}

From now on, assume that $\vec{x}$ and $\vec{y}$ are such that $\vec{x},\vec{y}\in \mathcal{X}_0$ and $\vec{y} = \vec{x}^*$.

By definition of $\mathcal{X}_0$, we have $\vec{d}^\top \nabla f(\vec{x}) = 0$, which together with  $\vec{u} + \vec{v} = \vec{x}^* - \vec{x}$, implies
\begin{equation}
\nabla f(\vec{x})^{\top} (\vec{x}^*-\vec{x}) = \nabla f(\vec{x})^{\top} \vec{v}.
\label{equ:e2}
\end{equation}

Now, note that for any $a\in [0,1]$, we have the following relations:

\begin{eqnarray*}
&& (\vec{x}^*-\vec{x})^{\top} \nabla^2 f(a \vec{x}^* + (1-a)\vec{x})(\vec{x}^*-\vec{x})\\
 &=& (\vec{u}+\vec{v})^{\top} \nabla^2 f(a \vec{x}^* + (1-a)\vec{x}) (\vec{u}+\vec{v})\\
&\overset{(a)}{\geq} & (\vec{u}+\vec{v})^{\top} \mat{A}_{\mathcal{X}} (\vec{u}+\vec{v})\\
&\overset{(b)}{\geq} & \vec{v}^{\top} \mat{A}_{\mathcal{X}} \vec{v}\\
&\geq & \left(\min_{\vec{y}: \vec{d}^\top \vec{y} = 0}\frac{\vec{y}^\top \vec{A}_{\mathcal{X}}\vec{y}}{||\vec{y}||^2}\right) ||\vec{v}||^2\\
& \geq & \gamma ||\vec{v}||^2
\end{eqnarray*}

where $(a)$ is by assumption (A1) and $(b)$ is by assumption that $\vec{A}_{\mathcal{X}}$ is a positive semidefinite matrix and (A2). Hence, we have shown that, for all $a\in [0,1]$,
\begin{equation}
(\vec{x}^*-\vec{x})^{\top} \nabla^2 f(a \vec{x}^* + (1-a)\vec{x})(\vec{x}^*-\vec{x}) \geq \gamma ||\vec{v}||^2.
\label{equ:e3}
\end{equation}

Next, note that
\begin{eqnarray*}
&& \nabla f(\vec{x})^{\top} \vec{v} + \frac{1}{2}\gamma ||\vec{v}||^2\\ 
&\geq & \min_{\vec{z}\in \reals^n} \left( \nabla f(\vec{x})^{\top} \vec{z} + \frac{1}{2} \gamma ||\vec{z}||^2 \right)\\
&\geq & -\frac{1}{2\gamma} ||\nabla f(\vec{x})||^2.
\end{eqnarray*}

Combining with (\ref{equ:e1})-(\ref{equ:e3}), we obtain
$$
f(\vec{x})-f(\vec{x}^*) \leq \frac{1}{2\gamma} ||\nabla f(\vec{x})||^2.
$$

\subsection{Proof of Lemma~\ref{lem:rhobounds}}\label{app:rhobounds}

\paragraph{Proof of (\ref{lem:ellpi})} Since $\ell(\pa) = \ell(\Pi(\pa))$ for all $\pa\in \reals^n$, we have that $\rho(\Pi(\pa))\geq \rho(\pa)$ is equivalent to $\ell_0(\Pi(\pa)) \geq \ell_0(\pa)$. 

Now, note
\begin{eqnarray*}
&& \ell_0(\Pi(\pa)) - \ell_0(\pa) \\
&=& \ell_0(\pa + c(\pa)\vec{1}) - \ell_0(\pa)\\
&=& (\alpha-1)n c(\pa) - \beta e^{c(\pa)}\sum_{i=1}^n e^{\pae_i} + \beta \sum_{i=1}^n e^{\pae_i}\\
&=& \beta \left(\sum_{i=1}^n e^{\pae_i}\right)\left(\frac{(\alpha-1)n}{\beta\sum_{i=1}^n e^{\pae_i}}c(\pa) - e^{c(\pa)} + 1\right)\\
&=& \beta \left(\sum_{i=1}^n e^{\pae_i}\right)e^{c(\pa)}\left(c(\pa)-1+e^{-c(\pa)}\right)\\
&\geq & 0
\end{eqnarray*}
where the last inequality holds by the fact $x - 1 + e^{-x}\geq 0$ for all $x\in\reals$.

\paragraph{Proof of (\ref{lem:ellort})} Indeed, $\nabla \rho (\pa) = \nabla \ell(\pa) + \nabla \ell_0(\pa)$. It is readily checked that $\nabla \ell(\pa)^{\top} \vec{1} = 0$ for all $\pa \in \reals^n$. We next show that $\nabla \ell_0(\Pi(\pa))^{\top} \vec{1} = 0$ for all $\pa \in \reals^n$. 

Note that 
$$
\frac{\partial}{\partial \pae_i}\ell_0(\pa) = \alpha -1 - \beta e^{\pae_i} \hbox{ for } i = 1,2,\ldots, n.
$$
Hence,
$$
\nabla \ell_0(\pa)^{\top} \mathbf{1} = (\alpha-1)n - \beta \sum_{i=1}^n e^{\pae_i}.
$$
Now, by definition of the mapping $\Pi$ given by (\ref{equ:map}) and (\ref{equ:cx}), for all $\pa\in \reals^n$,
$$
\nabla \ell_0(\Pi(\pa))^{\top} \vec{1} = (\alpha-1)n -\beta e^{c(\pa)}\sum_{i=1}^n e^{\pae_i} = 0.
$$

\subsection{Proof of Lemma~\ref{lm:raokupper}}\label{app:raokupper}

Let $t_{i,j}$ be the number of paired comparisons in the input data with tie outcome for items $i$ and $j$. Note that $t_{i,j} = t_{j,i}$. The log-likelihood function can be written as follows:
\begin{eqnarray*}
\ell(\pa) &=& \sum_{i=1}^n \sum_{j\neq i} \win_{i,j} \left(\pae_i - \log(e^{\pae_i} + \theta e^{\pae_j})\right) \\
&& + \frac{1}{2}\sum_{i=1}^n \sum_{j\neq i} t_{i,j}\left(\pae_i + \pae_j - \log(e^{\pae_i} + \theta e^{\pae_j})\right . \\
&& \left . - \log(\theta e^{\pae_i} + e^{\pae_j}) + \log(\theta^2-1)\right).
\end{eqnarray*}

Let $\bar{\win}_{i,j}$ be the number of paired comparisons of items $i$ and $j$ such that $i\succeq j$, i.e., $\bar{\win}_{i,j} = \win_{i,j} + t_{i,j}$. By a straightforward calculus, we can write
\begin{eqnarray*}
\ell(\pa) & = & \sum_{i=1}^n \sum_{j\neq i} \bar{\win}_{i,j} \left(\pae_i - \log(e^{\pae_i} + \theta e^{\pae_j})\right) \\
&& + \frac{1}{2}\sum_{i=1}^n t_{i,j} \log(\theta^2-1).
\end{eqnarray*}

Now, we note when $i\neq j$,
\begin{eqnarray*}
&& \frac{\partial^2}{\partial \pae_i \partial \pae_j} (-\ell(\pa))\\
& = & - \bar{\win}_{i,j} \frac{\theta e^{\pae_i}e^{\pae_j}}{(e^{\pae_i} + \theta e^{\pae_j})^2} - \bar{\win}_{j,i}\frac{\theta e^{\pae_i}e^{\pae_j}}{(\theta e^{\pae_i}+e^{\pae_j})^2}
\end{eqnarray*}
and
$$
\frac{\partial^2}{\partial \pae_i^2}(-\ell(\pa)) = -\sum_{j\neq i} \frac{\partial^2}{\partial \pae_u \partial \pae_j} (-\ell(\pa)).
$$

For any $i\neq j$, it indeed holds 
$$
\frac{\theta e^{\pae_i}e^{\pae_j}}{(e^{\pae_i} + \theta e^{\pae_j})^2} \leq \frac{1}{4}.
$$
Hence, when $i\neq j$,
$$
\frac{\partial^2}{\partial \pae_i \partial \pae_j} (-\ell(\pa)) \geq - \frac{1}{4}(\bar{\win}_{i,j} + \bar{\win}_{j,i}) \geq -\frac{1}{2}m_{i,j}.
$$

It follows that $\frac{1}{2}\vec{L}_\vec{M}\succeq \nabla^2 (-\ell(\pa))$ for all $\pa\in \reals^n$. Hence,
$$
\vec{x}^{\top} \nabla^2(-\ell(\vec{w}))\vec{x} \leq \frac{1}{2}\lambda_n(\vec{L}_\vec{M}) \hbox{ for all } \vec{x} \in \reals^n.
$$
This implies that $-\ell$ is a $\frac{1}{2}\lambda_n(\vec{L}_\vec{M})$-smooth function on $\reals^n$.

On the other hand, we can show that for all $\pa \in [-\bo,\bo]^n$,
$$
\frac{\theta e^{\pae_i}e^{\pae_j}}{(e^{\pae_i} + \theta e^{\pae_j})^2}  \geq \frac{\theta}{(\theta e^{-\bo} + e^{\bo})^2} := c_{\theta,\bo}.
$$
This can be noted as follows. Let $z = \theta e^{\pae_j}/(e^{\pae_i} + \theta e^{\pae_j})$. Note that
$$
\frac{\theta e^{\pae_i}e^{\pae_j}}{(e^{\pae_i} + \theta e^{\pae_j})^2} = z (1-z) 
$$
and that $ z\in \Omega := [1/(1+\theta e^{2\bo}), 1/(1 + \theta e^{-2\bo})]$. The function $z(1-z)$ is convex and thus achieves its minimum value over the interval $\Omega$ at one of its boundary points. It can be readily checked that the minimum is achieved at $z^* = 1/(1+\theta e^{2\bo})$, which yields $z^*(1-z^*) = c_{\theta,\bo}$.

Hence, when $i\neq j$,
$$
\frac{\partial^2}{\partial \pae_i \partial \pae_j} (-\ell(\pa)) \leq -c_{\theta,\bo} (\bar{\win}_{i,j} + \bar{\win}_{j,i})  \leq -c_{\theta,\bo}m_{i,j}.
$$

It follows that $\nabla^2 (-\ell(\pa)) \succeq c_{\theta,\bo} \vec{L}_\vec{M}$. From this, we have that for all $\pa \in [-\bo,\bo]^n$ and $\vec{x}\in \mathcal{X}$,
$$
\vec{x}^{\top} \nabla^2 (-\ell(\pa))\vec{x} \geq c_{\theta,\bo} \lambda_2(\vec{L}_\vec{M}) 
$$
where $\mathcal{X} = \{\vec{x}\in \reals^n: ||\vec{x}||_\infty \leq \bo \hbox{ and } \vec{x}^{\top}\vec{1} = 0\}$. This implies that $-\ell$ is $c_{\theta,\bo}\lambda_2(\vec{L}_\vec{M})$-strongly convex on $\mathcal{X}$.

\subsection{Proof of Lemma~\ref{lm:pl}}\label{app:pl}

It can be easily shown that for all $\pa\in [-\bo,\bo]^n$, $S\subseteq N$ such that $|S|\geq 2$, and $u,v\in S$ such that $u\neq v$, we have
$$
\frac{e^{-4\bo}}{|S|^2}\leq \frac{e^{\pae_u}e^{\pae_v}}{(\sum_{j\in S}e^{\pae_j})^2}\leq \frac{e^{4\bo}}{|S|^2}.
$$

Combining with (\ref{equ:BThessian}), we have
\begin{eqnarray*}
&& \frac{\partial^2}{\partial \pae_u\partial \pae_v}(-\ell(\pa)) \\
& \leq & - \sum_{y\in T}\win_y \frac{\pae_u \pae_v}{(\sum_{j=1}^k e^{\pae_{y_j}})^2}1_{u,v\in \{y_1,y_2,\ldots,y_k\}}\\
& \leq & - \frac{e^{-4\bo}}{k^2} \sum_{y\in T}\win_\pi  1_{u,v\in \{y_1,y_2,\ldots,y_k\}}\\
& = & - \frac{e^{-4\bo}}{k^2} m_{u,v}.
\end{eqnarray*}

From this it follows that for all $\vec{x}\in \reals^n$ such that $\vec{x}^{\top} \vec{1} = 0$, 
\begin{equation}
\vec{x}^{\top}\nabla^2 (-\ell(\pa))\vec{x} \geq \frac{e^{-4\bo}}{k^2}\lambda_2(\vec{L}_\vec{M}) ||\vec{x}||^2.
\label{equ:quad1}
\end{equation}

Similarly, we have
\begin{eqnarray*}
&& \frac{\partial^2}{\partial \pae_u \partial \pae_v}(-\ell(\pa)) \\
& \geq & - \sum_{y\in T}\win_y \sum_{l=1}^{k-1} \frac{\pae_u \pae_v}{(\sum_{j=l}^k e^{\pae_{y_j}})^2}1_{u,v\in \{y_1,y_2,\ldots,y_k\}}\\
& \geq & -e^{4\bo} \sum_{l=1}^{k-1}\frac{1}{(k-l+1)^2}\ m_{u,v}\\
&=& -e^{4\bo} \sum_{l=2}^{k}\frac{1}{l^2}\ m_{u,v}\\
&\geq & -e^{4\bo}\left(1+\int_1^k \frac{dx}{x^2}\right)m_{u,v}\\
&=& -e^{4\bo}\left(2-\frac{1}{k}\right)m_{u,v}.
\end{eqnarray*}

From this it follows that for all $\vec{x}$,
\begin{equation}
\vec{x}^{\top}\nabla^2 (-\ell(\pa))\vec{x} \leq e^{4\bo}\left(2-\frac{1}{k}\right)\lambda_n(\vec{L}_\vec{M}) ||\vec{x}||^2.
\label{equ:quad2}
\end{equation}

\subsection{Derivation of the convergence time bound (\ref{equ:unibound})}\label{app:unibound}

First note that 
$$
\underline{m} \vec{A}\leq \vec{M} \leq \bar{m} \vec{A}
$$ 
where the inequalities hold elementwise. From this, it follows that $\vec{L}_\vec{M} \succeq \underline{m}\vec{L}_\vec{A}$ and $\bar{m}\vec{L}_\vec{A} \succeq \vec{L}_\vec{M}$, where recall $\vec{A}$ is the adjacency matrix induced by matrix $\vec{M}$. Now, note
$$
\dM = ||\vec{M}||_\infty \leq \bar{m} d(n)
$$
and
$$
\aM = \lambda_2(\vec{L}_{\vec{M}})\geq \underline{m}\lambda_2(\vec{L}_{\vec{A}})
$$
where $d(n)$ is the maximum degree of a node in graph $G$.

Hence, we have
$$
\frac{\dM}{\aM} \leq \frac{r d(n)}{\lambda_2(\vec{L}_{\vec{A}})}.
$$

By Theorem 3.4 in \cite{M94}, for any graph $G$ with adjacency matrix $\vec{A}$ and diameter $D(n)$, $\lambda_2(\vec{L}_\vec{A})\geq 4/(nD(n))$. 

It thus follows that
$$
\frac{\dM}{\aM} \leq \frac{1}{4} r d(n) D(n) n
$$ 
which implies the convergence time bound $T = O(r d(n)D(n)n\log(1/\epsilon))$.

\subsection{Generalized Bradley-Terry models}
\label{sec:genBT-app}

In this section, we discuss how the results for Bradley-Terry model of paired comparisons can be extended to other instances of generalized Bradley-Terry models. In particular, we show this for the Rao-Kupper model of paired comparisons with tie outcomes, the Luce choice model and the Plackett-Luce ranking model. 

We discuss only the characterization of the strong-convexity and smoothness parameters as the convergence rate bounds for gradient descent and MM algorithms follow similarly as in Section~\ref{sec:conv-bt-gen}, from Theorems \ref{thm:conv-gd} and \ref{thm:conv-mm}, respectively. Similarly, the rate of convergence bounds for accelerated gradient descent and MM algorithms follow readily, similarly to as in Section~\ref{sec:accgen}, from Theorems, \ref{thm:accgd} and \ref{thm:accmm}, respectively.

\subsubsection{Model definitions}
\label{sec:genBT}

\paragraph{Bradley-Terry model of paired comparisons} According to the Bradley-Terry model, each paired comparison of items $i$ and $j$ has two possible outcomes: either $i$ wins against $j$ ($i\succ j$) or $j$ wins against $i$ ($j\succ i$). The distribution of the outcomes is given by  
$$
\Pr[i\succ j] = \frac{e^{\pae_i}}{e^{\pae_i} + e^{\pae_j}}
$$
where $\pa = (w_1, w_2,\ldots,w_n)^\top \in \reals^n$ are model parameters. 

\paragraph{Rao-Kupper model of paired comparisons with ties} The Rao-Kupper model is such that each paired comparison of items $i$ and $j$ has three possible outcomes: either $i\succ j$ or $j \succ i$ or $i\equiv j$ (tie). The model is defined by the probability distribution of outcomes that is given by
$$
\Pr[i\succ j] = \frac{e^{\pae_i}}{e^{\pae_i}+\theta e^{\pae_j}}
$$
and
$$
\Pr[i \equiv j] = \frac{(\theta^2-1)e^{\pae_i}e^{\pae_j}}{(e^{\pae_i} + \theta e^{\pae_j})(\theta e^{\pae_i} + e^{\pae_j})}
$$
where $\pa= (w_1, w_2,\ldots,w_n)^\top\in \reals^n$ and $\theta \geq 1$ are model parameters.

The larger the value of parameter $\theta$, the more mass is put on the tie outcome. For the value of parameter $\theta = 1$, the model corresponds to the Bradley-Terry model for paired comparisons.

\paragraph{Luce choice model} The Luce choice model is a natural generalization of the Bradley-Terry model of paired comparisons to comparison sets of two or more items. For any given comparison set $S\subseteq N=\{1,2,\dots,n\}$ of two or more items, the outcome is a choice of one item $i\in S$ (an event we denote as $i\succeq S$) which occurs with probability
$$
\Pr[i\succeq S] = \frac{e^{w_i}}{\sum_{j\in S} e^{w_j}}
$$
where $\vec{w}=(w_1, w_2,\ldots,w_n)^\top\in \reals_n$ are model parameters. 

We will use the following definitions and notation. Let $T$ be the set of ordered sequences of two or more items from $N$ such that for each $y = (y_1,y_2,\ldots,y_k) \in T$, $y_1$ is an arbitrary item and $y_2,\ldots, y_k$ are sorted in lexicographical order. We can interpret each $y = (y_1,y_2,\ldots,y_k) \in T$ as a choice of item $y_1$ from the set of items $\{y_1,y_2,\ldots,y_k\}$. According to the Luce's choice model, the probability of outcome $y$ is given by
$$
\Pr[Y = (y_1,y_2,\ldots,y_k)] = \frac{e^{\pae_{y_1}}}{\sum_{j\in y} e^{\pae_{j}}}. 
$$

We denote with $\win_y$ the number of observed outcomes $y$ in the input data. For each $y\in T$, let $|y|$ denote the number of items in $y$. 

\paragraph{Plackett-Luce ranking model} The Plackett-Luce ranking model is a model of full rankings: for each comparison set of items $S\subseteq N = \{1,2,\ldots,n\}$, the set of possible outcomes contains all possible permutations of items in $S$. The distribution over possible outcomes is defined as follows. Let $T$ be the set of all possible permutations of subsets of two or more items from $N$. Each $y = (y_1,y_2,\ldots,y_k) \in T$ corresponds to a permutation of the set of items $S = \{y_1,y_2,\ldots,y_k\}$. The probability of outcome $y$ is given by 
\begin{align*}
&\Pr[Y=(y_1, y_2,\ldots,y_k)] \\
= &\frac{e^{\pae_{y_1}}}{\sum_{j=1}^{k} e^{\pae_{y_j}}}
\frac{e^{\pae_{y_2}}}{\sum_{j=2}^{k} e^{\pae_{y_j}}}
\cdots
\frac{e^{\pae_{y_{k-1}}}}{\sum_{j=k-1}^{k} e^{\pae_{y_j}}}
\end{align*}
where $\pa = (w_1, w_2, \ldots,w_n)^\top \in \reals^n$ are model parameters. 

The model has an intuitive explanation as a sampling of items without replacement proportional to the item weights $e^{w_i}$. The Plackett-Luce ranking model corresponds to the Bradley-Terry model of paired comparisons when the comparison sets consist of two items. We denote with $\win_y$ the number of observed outcomes $y$ in the input data.

In this section, we discuss how the results for Bradley-Terry model of paired comparisons can be extended to other instances of generalized Bradley-Terry models. In particular, we show this for the Rao-Kupper model of paired comparisons with tie outcomes, the Luce choice model and the Plackett-Luce ranking model. 

\subsubsection{Rao-Kupper model} The probability distribution of outcomes according to the Rao-Kupper model is defined in Section~\ref{sec:genBT}. The log-likelihood function can be written as
\begin{eqnarray*}
\ell(\pa) &=& \sum_{i=1}^n \sum_{j\neq i} \bar{\win}_{i,j} \left(\pae_i - \log(e^{\pae_i}  + \theta e^{\pae_j})\right) \\
&& + \frac{1}{2}\sum_{i=1}^n t_{i,j} \log(\theta^2-1)
\end{eqnarray*}
where $\bar{\win}_{i,j}$ is the number of observed paired comparisons of items $i$ and $j$ such that either $i$ wins against $j$ or there is a tie outcome, and $t_{i,j}$ is the number of observed paired comparisons of items $i$ and $j$ with tie outcomes. 

\begin{lm} The negative log-likelihood function for the Rao-Kupper model of paired comparisons with parameter $\theta > 1$ is $\gamma$-strongly convex on $\mathcal{W}_\bo = \{\pa\in \reals^n : ||\pa||_\infty\leq \bo \hbox{ and } \pa^{\top} \vec{1} = 0\}$ and $\mu$-smooth on $\reals^n$ with
$$
\gamma = c_{\theta,\bo} \lambda_2(\mat{L}_{\mat{M}}) \hbox{ and } \mu = \frac{1}{2}\lambda_n(\mat{L}_{\mat{M}})
$$
where $c_{\theta,\bo}= \theta/(\theta e^{-\bo}+e^\bo)^2$.
\label{lm:raokupper}
\end{lm}

Proof of Lemma~\ref{lm:raokupper} is provided in Section~\ref{app:raokupper}.

A surrogate minorant function for the log-likelihood function of the Rao-Kupper model is given as follows:
\begin{eqnarray*}
&& \underline{\ell}(\vec{x}; \vec{y}) \\
&=& \sum_{i=1}^n \sum_{j\neq i} \bar{\win}_{i,j} \left(x_i -\frac{e^{x_i} + \theta e^{x_j}}{e^{y_i} + \theta e^{y_j}} - \log(e^{y_i} + \theta e^{y_j}) +1\right) \\
&& + \frac{1}{2}\sum_{i=1}^n t_{i,j} \log(\theta^2-1).
\end{eqnarray*}

The MM algorithm is defined by, for $i = 1,2,\ldots, n$,
\begin{align*}
&\pae_i^{(t+1)} = \log\left(\sum_{j\neq i} \bar{\win}_{i,j}\right) -\\
& \log\left(\sum_{j\neq i} \left(  \frac{\bar{\win}_{i,j} }{e^{\pae_{i}^{(t)}} + \theta e^{\pae_{j}^{(t)}}} +\frac{\theta \bar{\win}_{j,i} }{e^{\pae_{j}^{(t)}} + \theta e^{\pae_{i}^{(t)}}} \right)\right).
\end{align*}

\begin{lm} For all $\vec{x},\vec{y} \in [-\bo,\bo]^n$, $\underline{\ell}(\vec{x};\vec{y})-\ell(\vec{x})\geq -\frac{\delta}{2}||\vec{x}-\vec{y}||^2$ where
$$
\delta = e^{2\bo}\dM.
$$
\end{lm}

\subsubsection{Luce choice model} The probability distribution of outcomes according to the Luce choice model is defined in Section~\ref{sec:genBT}. The log-likelihood function can be written as:
$$
\ell(\pa) = \sum_{y\in T} \win_y \left(\pae_{y_1} - \log\left(\sum_{j\in y}e^{\pae_j}\right)\right).
$$

\begin{lm} The negative log-likelihood function for the Luce choice model with comparison sets of size $k\geq 2$ is $\gamma$-strongly convex and $\mu$-smooth on $\mathcal{W}_\bo = \{\pa\in \reals^n : ||\pa||_\infty\leq \bo \hbox{ and } \pa^{\top} \vec{1} = 0\}$ with
$$
\gamma = c_{\bo,k}\lambda_2(\mat{L}_\mat{M}) \hbox{ and } \mu = d_{\bo,k}\lambda_n(\mat{L}_\mat{M})
$$
where 
$$
c_{\bo,k} = \left\{
\begin{array}{ll}
1/(e^{-\bo}+e^\bo)^2, & \hbox{ if } k = 2\\
1/((k-2)e^{2\bo}+2)^2, & \hbox{ if } k > 2
\end{array}
\right .
$$
and
$$
d_{\bo,k} = \frac{1}{((k-2)e^{-2\bo}+2)^2}.
$$
\end{lm}

Note that for every fixed $\bo > 0$, (a) $c_{\bo,k}/d_{\bo,k}$ is decreasing in $k$, (b) $1/e^{8\bo}\leq c_{\bo,k}/d_{\bo,k}\leq 1/e^{2\bo}$, and (c) $1/e^{8\bo}$ is the limit value of $c_{\bo,k}/d_{\bo,k}$ as $k$ goes to infinity.

A minorant surrogate function for the log-likelihood function of the Luce choice model is given by
$$
\underline{\ell}(\vec{x}; \vec{y}) = \sum_{y\in T} \win_y \left(x_{y_1} - \frac{\sum_{j\in y} e^{x_j}}{\sum_{j\in y} e^{y_j}} - \log\left(\sum_{j\in y} e^{y_j}\right) + 1\right).
$$

The MM algorithm iteration can be written as: for $i=1,2,\ldots,n$, 
\begin{eqnarray*}
\pae_i^{(t+1)} &=& \log\left(\sum_{y\in T} \win_y \ind_{i = y_1}\right) \\
&& - \log\left(\sum_{y\in T} \win_y \ind_{i\in y}\frac{1}{\sum_{j\in y} e^{\pae_{j}^{(t)}}}\right)
\end{eqnarray*}
where $\sum_{y\in T} \win_y \ind_{i = y_1}$ is the number of observed comparisons in which item $i$ is the chosen item.

\begin{lm} For all $\vec{x},\vec{y} \in [-\bo,\bo]^n$, $\underline{\ell}(\vec{x};\vec{y})-\ell(\vec{x})\geq -\frac{\delta}{2}||\vec{x}-\vec{y}||^2$ where
$$
\delta = \frac{1}{k(k-1)}e^{2\bo}\dM.
$$
\end{lm}

\subsubsection{Plackett-Luce ranking model} The probability distribution of outcomes according to the Plackett-Luce ranking model is defined in Section~\ref{sec:genBT}. The log-likelihood function can be written as follows:
$$
\ell(\pa) = \sum_{y\in T} \win_y \sum_{r=1}^{|y|-1} \left( \pae_{y_r} - \log\left(\sum_{j=r}^{|y|} e^{\pae_{y_{j}}}\right)\right).
$$

\begin{lm} The negative log-likelihood function for the Plackett-Luce ranking model with comparison sets of size $k\geq 2$ is $\gamma$-strongly convex and $\mu$-smooth on $\mathcal{W}_\bo = \{\pa\in \reals^n : ||\pa||_\infty\leq \bo \hbox{ and } \pa^{\top} \vec{1} = 0\}$ with
$$
\gamma = \tilde{c}_{\bo,k}\lambda_2(\mat{L}_\mat{M}) \hbox{ and } \mu = \tilde{d}_{\bo,k}\lambda_n(\mat{L}_\mat{M})
$$
where 
$$
\tilde{c}_{\bo,k} = \frac{1}{k^2}e^{-4\bo}
\hbox{ and }
\tilde{d}_{\bo,k} = \left(2-\frac{1}{k}\right)e^{4\bo}.
$$
\label{lm:pl}
\end{lm}

Proof of Lemma~\ref{lm:pl} is provided in Section~\ref{app:pl}.

Note that for fixed values of $\bo$ and $k$, Lemma~\ref{lm:pl} implies the convergence time $\log(\dM/\aM)$. Note, however, that for fixed $\bo > 0$, $\tilde{c}_{\bo,k}/\tilde{d}_{\bo,k}$ decreases to $0$ with $k$ and is of the order $1/k^2$. This is because in the derivation of parameters $\tilde{c}_{\bo,k}$ and $\tilde{d}_{\bo,k}$ we use (conservative) deterministic bounds.  Following \cite{hajek14}, one can derive bounds for $\gamma$ and $\mu$ that hold with high probability, which are such that $\tilde{c}_{\bo,k}$ and $\tilde{d}_{\bo,k}$ scale with $k$ in the same way.

The log-likelihood function of the Plackett-Luce ranking model admits the following minorization function:
\begin{small}
\begin{eqnarray*}
&& \underline{\ell}(\vec{x}; \vec{y}) \\
&=& \sum_{y\in T} \win_y \sum_{r=1}^{|y|-1} \left(x_{y_r} - \frac{\sum_{j=r}^{|y|}e^{x_{y_{j}}}}{\sum_{j=r}^{|y|}e^{y_{y_{j}}}} - \log\left(\sum_{j=r}^{|y|}e^{y_{y_{j}}}\right) + 1\right).
\end{eqnarray*}
\end{small}

The MM algorithm is given by: for $i = 1,2,\ldots,n$,
\begin{eqnarray*}
\pae_i^{(t+1)} &=& \log\left(\sum_{y\in T} \win_y \ind_{i\in S_{1,|y|-1}(y)}\right) \\
&& - \log\left(\sum_{y\in T} \win_y \sum_{r=1}^{|y|-1}\ind_{i\in S_{r,|y|}(y)} \frac{1}{\sum_{j=r}^{|y|} e^{\pae_{y_j}^{(t)}}}\right)
\end{eqnarray*}
where $S_{a,b}(y) = \{y_a, y_{a+1},\ldots, y_b\}$.

\begin{lm} For all $\vec{x},\vec{y} \in [-\bo,\bo]^n$, $\underline{\ell}(\vec{x};\vec{y})-\ell(\vec{x})\geq -\frac{\delta}{2}||\vec{x}-\vec{y}||^2$ where
$$
\delta = \frac{1}{2}e^{2\bo}\dM.
$$
\end{lm}



\bibliographystyle{informs2014}
\bibliography{ref}

\end{document}